\documentclass{article}
\usepackage[utf8]{inputenc}

\usepackage[nonatbib, final]{neurips_2023}

\usepackage{graphicx}
\usepackage{xcolor}
\usepackage{amsmath}
\usepackage{amssymb}
\usepackage{amsfonts}
\usepackage{bm}
\usepackage{booktabs}
\usepackage{chngcntr}
\usepackage[group-separator={,}]{siunitx}

\usepackage[square,numbers]{natbib}
\usepackage{color}
\definecolor{mydarkblue}{rgb}{0,0.08,0.45}
\usepackage[colorlinks=true,allcolors=mydarkblue]{hyperref}

\definecolor{gray40}{gray}{0.4}
\newcommand{\tablehelper}[1]{\textcolor{gray40}{\tiny ($\pm {#1}$)}}
\newcommand{\plusminus}[1]{\textcolor{gray40}{\tiny $\pm {#1}$}}
\newcommand{\emptyhelper}[1]{\textcolor{gray40}{}}

\usepackage{wrapfig}

\newcommand{\E}{\mathbb{E}}
\newcommand{\N}{\mathcal{N}}

\newcommand{\KL}{\text{KL}}

\newcommand{\PoE}{\text{PoE}}
\newcommand{\MoE}{\text{MoE}}
\newcommand{\MoPoE}{\text{MoPoE}}

\newcommand{\boldI}{\mathbf{I}}

\newcommand{\boldy}{\mathbf{y}}
\newcommand{\boldx}{\mathbf{x}}

\newcommand{\boldz}{\mathbf{z}}

\newcommand{\GP}{\mathcal{GP}}

\newcommand{\boldzero}{\mathbf{0}}

\newcommand{\calI}{\mathcal{I}}

\newcommand{\calM}{\mathcal{M}}
\newcommand{\calL}{\mathcal{L}}
\newcommand{\calP}{\mathcal{P}}

\newcommand{\calX}{\mathcal{X}}
\newcommand{\calZ}{\mathcal{Z}}

\newcommand{\zshared}{\boldz^{\text{shared}}}
\newcommand{\zprivateone}{\boldz^{\text{pr}_1}}
\newcommand{\zprivatetwo}{\boldz^{\text{pr}_2}}

\title{Disentangling shared and private latent factors in multimodal Variational Autoencoders}

\author{%
  Kaspar Märtens \\
  University of Oxford
  \And
  Christopher Yau \\
  University of Oxford \\
  The Alan Turing Institute \\
  Health Data Research UK
}

\begin{document}

\setlength{\abovedisplayskip}{3pt}
\setlength{\belowdisplayskip}{3pt}

\maketitle

\begin{abstract}
    Generative models for multimodal data permit the identification of latent factors that may be associated with important determinants of observed data heterogeneity. Common or shared factors could be important for explaining variation across modalities whereas other factors may be private and important only for the explanation of a single modality. Multimodal Variational Autoencoders, such as MVAE and MMVAE, are a natural choice for inferring those underlying latent factors and separating shared variation from private. In this work, we investigate their capability to reliably perform this disentanglement. In particular, we highlight a challenging problem setting where modality-specific variation dominates the shared signal. Taking a cross-modal prediction perspective, we demonstrate limitations of existing models, and propose a modification how to make them more robust to modality-specific variation. Our findings are supported by experiments on synthetic as well as various real-world multi-omics data sets. 
\end{abstract}

\section{Introduction}

Multimodal data integration is an important task for increasingly data rich applications. In biology, the integration of data from different molecular assays enables greater understanding of biological processes and disease mechanisms. These \emph{multi-omic} profiling studies that probe different biological layers in parallel have become increasingly common \citep{stuart_integrative_2019, hasin_multi-omics_2017, krassowski_state_2020}. 
Multi-omic data analysis is typically concerned with identification of \textit{latent factors} from the combination of these high-dimensional data modalities (also referred to as \emph{views}). These latent factors are assumed to be associated with important physical drivers of biological heterogeneity. However, this analysis is complicated by the complex dependencies across modalities that must be inferred as well as the variable dimensionality due to the different number and types of features measured.

In complex systems, it can be expected that latent processes segregate into two groups, those that drive variation across all modalities (shared factors), and those that only influence variation in a single modality (private factors). 
% Common or shared factors could be important for explaining variation across modalities whereas other factors may be private and important only for the explanation of a single modality. 
Disentangling shared and private variation can therefore be important for interpretation but may be difficult as different feature sets could be driven by different latent factors (Figure 1A). Historically, finding shared signal across two data modalities has been performed with canonical correlation analysis (CCA)  \citep{hotelling_relations_1936}. 
Inter-battery factor analysis (IBFA) \citep{tucker_inter-battery_1958} can be seen as an extension of CCA that additionally incorporates \emph{modality-specific} latent factors. Bayesian formulations of CCA and IBFA have been developed using generative modelling \citep{klami_generative_2006, ek_shared_2009, virtanen_bayesian_2012, klami_bayesian_2013}. 
These have also been extended to non-linear setups by, for example, combining CCA with deep neural networks \citep{wang_deep_2016, gundersen_end--end_2020}, or using Gaussian Process Latent Variable Models for multi-view learning \citep{damianou_multi-view_2021}. 

Generative models are a natural choice for the task of learning those underlying latent factors. In particular, Variational Autoencoders (VAEs) \citep{kingma_auto-encoding_2014, rezende_stochastic_2014} have been successfully used in various biological applications \citep{lopez_deep_2018, lotfollahi_predicting_2023, weinberger_disentangling_2022}, and their multimodal adaptations have been used for multi-omics data integration \citep{minoura_mixture--experts_2021,gayoso_joint_2021,lotfollahi_multigrate_2022,cao_integrated_2022,liu_cvqvae_2022}. 
However, while these works have showcased the capabilities of multimodal VAEs in representation learning, few attempt to combine this with latent factor disentanglement to improve their interpretability.

% VAEs have been extended to multiple data modalities in a few different ways. 
In multimodal VAEs, the \emph{encoder} projects observations onto distributions in the latent space. One of the main challenges is the design choice of the encoder and the form of the associated approximate posterior. Specifically, in order to allow for missing modalities at either training or test time, typical approaches introduce modality-specific encoders and then combine them in different ways. 

Two prominent and widely-used approaches to multimodal VAEs are the MVAE \citep{wu_multimodal_2018} and the MMVAE \citep{shi_variational_2019}. MVAE uses a product-of-experts (PoE) to combine the encoding distributions, whereas the MMVAE uses a mixture-of-experts (MoE). 
%From the perspective of approximate inference, the PoE and MoE are simply two
Both approaches have been adopted in various multi-omics applications, including \citep{lee_variational_2021, lotfollahi_multigrate_2022} (PoE) and \citep{minoura_mixture--experts_2021} (MoE). More recently, a generalisation based on the mixture-of-product-of-experts (MoPoE) has been proposed \citep{sutter_generalized_2021}, but has not yet been used in computational biology applications. Recently both MVAE and MMVAE frameworks have been adapted to explicitly include shared and private latent variables \citep{lee_private-shared_2021, palumbo_mmvae_2022} to enable the separation of modality-specific variation from shared and therefore improved interpretation as considered in various biomedical applications, e.g.\ as considered in \citep{argelaguet_multi-omics_2018, gundersen_end--end_2020, ash_joint_2021}. However, PoE and MoE are distinct choices for the form of the variational posterior. 
One is not a special case of the other, so it is unclear if both approaches offer the same capability to disentangle variation nor have their resilience to variable dimensionality been demonstrated. The latter drives the relative proportions of modality-specific and shared variation, e.g. gene expression involves measuring $O(10^3-10^4)$ of genes but DNA methylation measures $O(10^5-10^6)$ CpG sites.  

\begin{figure}[!t]
    \centering
    \includegraphics[width=\columnwidth]{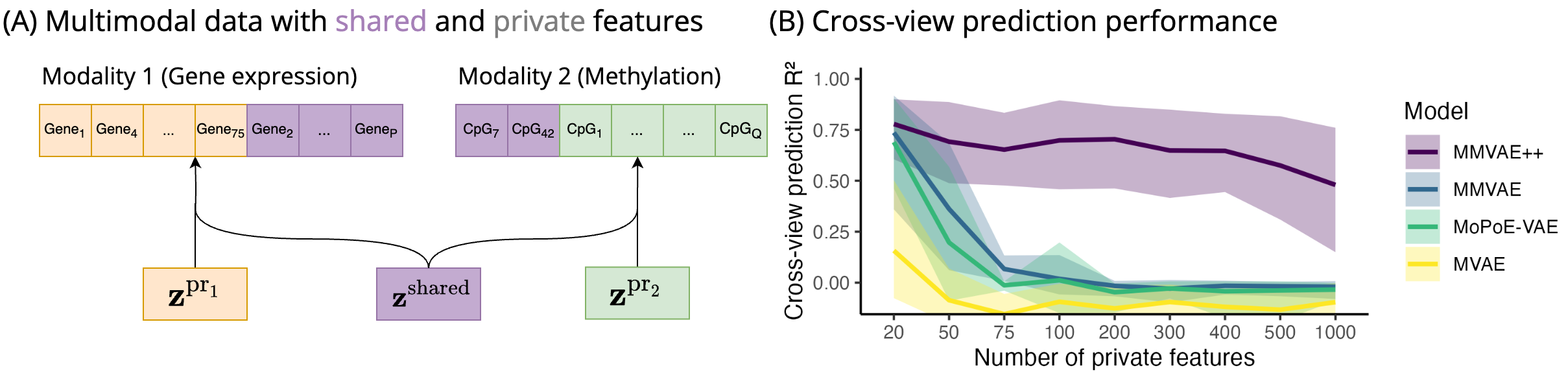}
    \vspace{-1.5em}
    \caption{
        \textbf{(A) Multimodal data with the underlying shared and private latent variables.} We consider multimodal data (illustrated for tabular omics data, such as gene expression and methylation), where we distinguish between features that have been generated by \emph{shared} latent factors and features that are driven by \emph{private} modality-specific sources of variation.
        \textbf{(B) Cross-modal prediction performance when increasing the number of private features.} When gradually increasing the number of features that are driven by private latent factors (based on the example from Section~\ref{sec:synthetic}), we observe that: 1) MMVAE and MoPoE consistently outperform MVAE, 2) the performance of all methods drops when increasing private features, but 3) our proposed modification MMVAE++ is significantly more robust than existing methods to high modality-specific variation.  %in identifying shared latent factors which are required for successful cross-modal prediction. 
    }
    \vspace{-1em}
    \label{fig:figure1}
\end{figure}

\textbf{Contributions:} We investigate the ability of multimodal VAEs to successfully disentangle \emph{private} and \emph{shared} sources of variation in multi-omics data. We highlight a challenging real-world scenario where dominant modality-specific variation can make it harder to identify the more subtle shared latent structure. We will demonstrate that the existing approaches (MVAE, MoPoE-VAE and MMVAE) exhibit systematic differences in their ability to identify the underlying shared latent factors, stemming from the \emph{inductive biases} associated with the use of MoE/PoE posterior approximations. We explain this through the lens of cross-modal prediction. %their role in cross-modal prediction /  and their relation to cross-modal prediction
Our experiments highlight how all models behave desirably when there is relatively little modality-specific variation, but they start to break down when private variation starts to dominate. To address this shortcoming, we further propose a modification called MMVAE++ that is significantly more robust in its ability to infer shared latent factors.

\section{Background}

\subsection{Variational Autoencoders}

VAEs are a class of deep generative models that couple a Bayesian latent variable model with an inference network. 
Given observed data $\boldx \in \calX$, the goal of latent variable modelling is to associate data $\boldx$ with a latent representation $\boldz \in \calZ$, where typically $\dim(\boldz) \ll \dim(\boldx)$. 
VAEs specify a joint distribution $p(\boldx, \boldz) = p_{\theta}(\boldx |\boldz) p(\boldz)$, where the likelihood $p_{\theta}(\boldx |\boldz)$ is parameterised by a \emph{decoder} neural network with parameters $\theta$, and the prior is typically chosen to be $p(\boldz) = \N(\boldzero, \boldI)$. 
Posterior inference in VAEs is carried out via \emph{amortised} variational inference, i.e.\ using a parametric inference model $q_\phi(\boldz|\boldx)$. This is implemented via an \emph{encoder} neural network with parameters $\phi$, a typical choice being $q_\phi(\boldz|\boldx) = \N(\mu_\phi(\boldx), \sigma^2_\phi(\boldx))$. 
Application of standard variational inference methodology  \citep{blei_variational_2017} leads to a lower bound on the log marginal likelihood, i.e.\ the evidence lower bound (ELBO) of the form $\mathcal{L} = \mathbb{E}_{q_{\phi}(\boldz | \boldx)} \log p_\theta(\boldx | \boldz)
    - \text{KL}(q_{\phi}(\boldz|\boldx) || p(\boldz))$. 
% \begin{align*}
%     \mathcal{L} = \mathbb{E}_{q_{\phi}(\boldz | \boldx)} \log p_\theta(\boldx | \boldz)
%     - \text{KL}(q_{\phi}(\boldz|\boldx) || p(\boldz)) .
% \end{align*}
VAEs are trained by optimising $\calL$ w.r.t.\ both the decoder parameters $\theta$ and variational parameters $\phi$.

\subsection{Multimodal VAEs}

In principle, VAEs can be extended to multiple data modalities $\boldx_{1:M}$ in a relatively straightforward way by constructing the decoder $p(\boldx_1, \ldots, \boldx_M | \boldz)$ and encoder $q(\boldz | \boldx_1, \ldots, \boldx_M)$, now involving all data modalities. 
For example, this can be done via concatenation
% \footnote{Inputs to the encoder could either be concatenated directly, or they could first be transformed so one can concatenate their modality-specific representations.}
as in JMVAE \citep{suzuki_joint_2016}. 
However, this type of approach is limited to complete data situations as there is no principled mechanism to support missing data modalities at train and test time otherwise.  
To construct a multimodal VAE that can handle missing views, a natural approach is to introduce modality-specific encoders $q_{\phi_m}(\boldz | \boldx_m)$ for $m = 1, \ldots, M$. 
These can then be combined for the modalities that are available via either a product of experts (PoE), a mixture of experts (MoE), or a mixture of PoEs (MoPoE) leading to three different implementation variants of multimodal VAEs.

%Here the key question is how to combine these modality-specific encoding distributions. 

%. There are two main approaches for how these encoders can be combined: \citet{wu_multimodal_2018} propose to combine the encoders via a product of experts (PoE), whereas \citet{shi_variational_2019} propose to use a mixture of experts (MoE). The two respective models are referred to as MVAE and MMVAE. We will now review both in more detail. 

% There is more than one way to extend VAEs to multiple data modalities $\boldx_{1:M}$. 

\paragraph{MVAEs:}

The MVAE \citep{wu_multimodal_2018} uses a product of experts (PoE) \citep{hinton_training_2002} to combine the variational posterior approximations $q_{\phi_m}(\boldz |\boldx_m)$. This choice is motivated by how the true posterior in the multimodal VAE factorises. For a set of modalities $\boldx_\calM$, where $\calM$ is an index set, MVAE uses the PoE including a ``prior expert'' $p(\boldz)$ as follows
\begin{align*}
    q^\PoE_\phi(\boldz | \boldx_\mathcal{M}) := p(\boldz) \prod_{m \in \mathcal{M}} q_{\phi_m}(\boldz |\boldx_m)
\end{align*}
% and the corresponding ELBO $\mathcal{L}_{\mathcal{M}}^\PoE := 
%    \E_{\boldz \sim q^{\PoE}(\boldz | \boldx_{\calM})} \log p(\boldx_{\calM} | \boldz) - \KL(q_\phi^{\PoE}(\boldz | \boldx_{\mathcal{M}}) \,||\, p(\boldz))$
% \begin{align} \label{eq:MVAE_ELBO}
%     \mathcal{L}_{\mathcal{M}}^\PoE := 
%     \E_{\boldz \sim q^{\PoE}(\boldz | \boldx_{\calM})} \log p(\boldx_{\calM} | \boldz) - \KL(q_\phi^{\PoE}(\boldz | \boldx_{\mathcal{M}}) \,||\, p(\boldz))
% \end{align}
% \begin{align} \label{eq:MVAE_ELBO}
%     \mathcal{L}_{\mathcal{M}}^\PoE := 
%     \E_{\boldz \sim q^{\PoE}(\boldz | \boldx_{\calM})} \log p(\boldx_{\calM} | \boldz) \\ \nonumber - \KL(q_\phi^{\PoE}(\boldz | \boldx_{\mathcal{M}}) \,||\, p(\boldz))
% \end{align}
where typically all experts are chosen to be Gaussian. 
%evaluated on an index set $\mathcal{M}$. 
While it would appear sufficient to optimise the corresponding ELBO, which we denote $\calL^{\PoE}_{\calM}$, calculated across the set of all modalities, i.e.\ the objective $\calL_{{1:M}}^\PoE$, there turns out to be a limitation. 
The product-of-Gaussians does not uniquely specify its components, so individual inference networks would not be uniquely defined. This would be problematic when using them at test time in the presence of missing data. 
To address this issue, the authors \citet{wu_multimodal_2018} propose to optimise a sum of ELBOs, considering different index sets $\calM$. In case of $M=2$ views, these index sets would be $\{1, 2\}, \{1\}$ and $\{2\}$, thus resulting in the following MVAE objective
\begin{align} \label{eq:MVAE_sum_of_ELBOs}
    \mathcal{L}^\text{MVAE} := \mathcal{L}_{\{1, 2\}}^\PoE + \mathcal{L}_{\{1\}}^\PoE + \mathcal{L}_{\{2\}}^\PoE \, .
\end{align}
% We refer to the above as the MVAE objective. 

\paragraph{MMVAEs:}

In contrast to using the PoE, the MMVAE by \citet{shi_variational_2019} adopts a mixture of experts (MoE) \citep{jacobs_adaptive_1991} variational posterior, where the modality-specific encoders $q_{\phi_m}(\boldz |\boldx_m)$ are combined additively $q^\MoE_\phi (\boldz | \boldx_\calM) = \sum_{m \in \calM} \alpha_m q_{\phi_m}(\boldz |\boldx_m)$. 
% \begin{align*}
%     q^\MoE_\phi (\boldz | \boldx_\calM) = \sum_{m \in \calM} \alpha_m q_{\phi_m}(\boldz |\boldx_m)
% \end{align*}
Here, \citet{shi_variational_2019} choose equal weights $\alpha_m = 1/M$.

\paragraph{MoPoE-VAEs:}

\citet{sutter_generalized_2021} introduce a mixture-of-product-of-experts (MoPoE) posterior approximation, $q^\MoPoE_\phi (\boldz | \boldx_{1:M}) \propto \sum_{\calM \in \calP(1:M)} q^\PoE_\phi(\boldz | \boldx_\mathcal{M})$
% \begin{align*}
%     q^\MoPoE_\phi (\boldz | \boldx_{1:M}) = \frac{1}{2^M} \sum_{\calM \in \calP(1:M)} q^\PoE_\phi(\boldz | \boldx_\mathcal{M}) % \prod_{m \in \calM} q_{\phi_m}(\boldz |\boldx_m)
% \end{align*}
where the mixture is taken over all subsets of modalities. This can be seen as a generalisation of both the MoE and the PoE.  

\subsection{Private and shared latent factors in multimodal VAEs} \label{sec:private_and_shared}

Recently, the MVAE and MMVAE frameworks have been adapted to explicitly include shared and private latent variables by \citet{lee_private-shared_2021} and \citet{palumbo_mmvae_2022} respectively. 
Both of these papers essentially introduce a partitioned latent space, in case of two modalities $\boldz = [\zprivateone, \zshared, \zprivatetwo]$ as illustrated in Figure~\ref{fig:figure1}A\footnote{While both papers come with their own additional minor modifications, for example \citet{palumbo_mmvae_2022} replace the Gaussian prior with a Laplace one. For a consistent comparison, here we consider the ``base'' versions of both models, i.e\ apart from the partitioned latent space, we leave other aspects of the model specification unchanged.}. 
This involves learning $q_{\psi_{m}^1}(\boldz_{\text{pr}_1} |\boldx_m), q_{\psi_{m}^2}(\boldz_{\text{pr}_2} |\boldx_m)$ and $q_{\phi_m}(\boldz_{\text{sh}} |\boldx_m)$ encoding distributions.

\section{Methods}

We proceed by reinterpreting the MVAE, MMVAE, and MoPoE-VAE models from the cross-modal generation perspective. %, seeking to explain the empirical findings from Figure~\ref{fig:figure1}.
In our setting, the effectiveness of the latent disentanglement into private and shared components can be measured by the ability to predict one data modality from another. Since cross-modal prediction can only be achieved through the \textit{shared} latent space, accurate predictions can only be achieved if the learnt shared components contain only truly shared information and avoids modality-specific information.

\subsection{Inductive biases in existing multimodal VAEs: a cross-modal perspective} \label{sec:inductive_biases}

Cross-modal prediction refers to predicting one set of modalities $\boldx_\calM$ from a different set of modalities $\boldx_\calI$. In multimodal VAEs, this is achieved as follows
\begin{align} \label{eq:crossmodal_pred}
    p(\boldx_\calM | \boldx_\calI) \approx \int p(\boldx_\calM | \boldz) q(\boldz | \boldx_\calI) d \boldz \, 
\end{align}
where we first encode $\boldx_\calI$ and then decode to obtain  predictions for $\boldx_\calM$. 
Next, we will discuss how the MVAE, MMVAE, and MoPoE-VAE training objectives relate to cross-modal prediction.

\paragraph{PoE vs MoE on fixed inputs:} 
Usually, the PoE and the MoE are interpreted as simply two different choices for the variational posterior $q(\boldz | \boldx_{1:M})$. To an extent, the differences between PoE and MoE are well understood. For example, PoE will have high posterior density only as long as \emph{all} experts have high posterior density, whereas in MoE it is sufficient that there exists \emph{at least one} expert that has assigned high posterior density to this event. 
While these differences between MVAE and MMVAE are well-understood, the above reasoning assumes that we compare the two on the \emph{same set of inputs}, but this is not necessarily the case when performing inference for latent variables.

\paragraph{MVAE vs MMVAE:}
The MVAE objective does not involve any cross-modal reconstruction terms. In contrast, the MMVAE lower bound, given by $\E_{\boldz \sim q^{\MoE}(\boldz | \boldx_{\calM})} \log p(\boldx_{\calM} | \boldz) - \KL(q_\phi^{\MoE}(\boldz | \boldx_{\mathcal{M}}) \,||\, p(\boldz))$, where the first term can be expanded 
\begin{align} \label{eq:MoE_all_pairs}
    \frac{1}{M} \sum_{m \in \calM} \sum_{m' \in \calM} \E_{\boldz \sim q_{\phi_m}(\boldz | \boldx_m)} \log p(\boldx_{m'} | \boldz) \, ,
\end{align}
involves a sum over all pairs $(m, m')$ of modalities ($1 \le m, m' \le M$),
thus explicitly involving cross-modal reconstruction. 
We hypothesise that this is an important distinction which may lead to MMVAE learning \emph{latent representations that are more amenable to cross-modal prediction}, i.e.\ being potentially biased towards learning \emph{shared} latent factors, more so than MVAE. This inductive bias may be particularly important in the abundance of modality-specific variation. In this scenario (e.g.\ as in Figure~\ref{fig:figure1}A), the MVAE may absorb this dominant, private source of variation in the shared latent space $\zshared$, whereas the MMVAE objective may correctly capture the non-dominant, but \emph{shared} variation in $\zshared$.

\paragraph{MVAE vs MoPoE-VAE:}
Finally, we aim to connect and contrast MVAE and MoPoE-VAE. 
In the original MVAE objective $\calL^\PoE_{\calM}$, data $\boldx_\calM$ is first encoded into the latent space $\boldz$ and then decoded to reconstruct $\boldx_\calM$. While it seems natural that both \emph{encoding} and \emph{decoding} are performed on the same set of modalities $\boldx_{\calM}$, this does not necessarily have to be the case. 
Inspired by the above~\eqref{eq:crossmodal_pred}, we introduce a more general notation, now explicitly indicating the \emph{encoding} index set $\calI$
\begin{align*}
    \calL_{\calM \leftarrow \calI} := 
    \E_{\boldz \sim q^{\PoE}(\boldz | \boldx_{\calI})} \log p(\boldx_{\calM} | \boldz) - \KL(q^{\PoE}(\boldz | \boldx_{\mathcal{I}}) \,||\, p(\boldz)) \,
\end{align*}
which is a valid lower bound (this can be seen as choosing an approximate posterior $q^{\PoE}(\boldz | \boldx_{\calI})$ which is amortised w.r.t.\ inputs $\boldx_{\calI}$). 
Using this notation, in case of $M=2$ modalities, the MVAE lower bound can be written as $\mathcal{L}_{\{1, 2\} \leftarrow \{1, 2\}} + \mathcal{L}_{\{1\} \leftarrow \{1\}} + \mathcal{L}_{\{2\} \leftarrow \{2\}}$, letting us directly compare it with the MoPoE-VAE objective
\begin{align} \label{eq:MoPoE}
    \mathcal{L}_{\{1, 2\} \leftarrow \{1, 2\}} + 
    \mathcal{L}_{\{1\} \leftarrow \{1\}} + 
    \mathcal{L}_{\{2\} \leftarrow \{2\}} + 
    \mathcal{L}_{\{1\} \leftarrow \{2\}} + 
    \mathcal{L}_{\{2\} \leftarrow \{1\}} \, .
\end{align}
The fact that the last two cross-modal terms exist in MoPoE-VAE but not in MVAE, helps to shed light into the expected behaviours of MVAE and MoPoE-VAE, specifically on why we would expect MoPoE-VAE to be better suited for cross-modal reconstruction than MVAE. 
We also note that, in principle, one could have constructed a different version of MVAE, considering all possible configurations of encoding and decoding modalities $\calI$ and $\calM$. This would have lead to inclusion of cross-modal terms $\mathcal{L}_{\{1\} \leftarrow \{2\}}$ and $\mathcal{L}_{\{2\} \leftarrow \{1\}}$.

\subsection{MMVAE++}

We now turn to MMVAE. In Section~\ref{sec:inductive_biases} we saw why MMVAE is better suited for cross-view prediction than MVAE. However, as we see in Figure~\ref{fig:figure1}B, it is still prone to the presence of relatively high levels of modality-specific variation. As this is a common occurrence in real-world data due to the variable dimensionality of different modalities, it would be highly desirable to improve the robustness of MMVAE in these situations. 

Modality-specific latent structure can interfere with learning \emph{shared} latent factors. This can happen because being able to accurately reconstruct a large number of private features can lead to a higher ELBO, in comparison to reconstructing a relatively smaller number of shared features. 
% In the MMVAE objective, we can separate same-view and cross-view reconstructions. 
Even when explicitly separating out shared and private latent variables, i.e.\ when assuming a latent space $\boldz := [\zprivateone, \zshared, \zprivatetwo]$, it is not guaranteed that $\zshared$ will contain latent variables that are \emph{truly shared}. 
In order to enforce this, we propose a modification to the MMVAE objective, where the gradient updates via $\zshared$ are restricted to only cross-modal terms. That is, we redefine
%is optimised only using cross-modal terms, as follows:
\begin{align*}
\E{}_{\boldz \sim q(\boldz | \boldx_m)} \log p(\boldx_{m'} | \boldz) 
    =
    \begin{cases}
    \E_{\substack{
      \boldz_\text{pr} \sim q(\boldz_\text{pr} | \boldx_m) \\
      \boldz_\text{sh} \sim \text{sg}[q(\boldz_{\text{sh}} | \boldx_m)]
    }} \log p(\boldx_{m'} | \boldz_{\text{sh}}, \boldz_{\text{pr}}),  \text{ if } m = m',
    \\
    \E_{\substack{
      \boldz_\text{sh} \sim q(\boldz_{\text{sh}} | \boldx_m)
    }} \log p(\boldx_{m'} | \boldz_{\text{sh}}), \text{ if } m \ne m',
    \end{cases}
\end{align*}
% \begin{align*}
% \E{&}_{\boldz \sim q(\boldz | \boldx_m)} \log p(\boldx_{m'} | \boldz) \\
%     &=
%     \begin{cases}
%     \E_{\substack{
%       \boldz_\text{pr} \sim q(\boldz_\text{pr} | \boldx_m) \\
%       \boldz_\text{sh} \sim \text{sg}[q(\boldz_{\text{sh}} | \boldx_m)]
%     }} \log p(\boldx_{m'} | \boldz_{\text{sh}}, \boldz_{\text{pr}}),  \text{ if } m = m',
%     \\
%     \E_{\substack{
%       \boldz_\text{sh} \sim q(\boldz_{\text{sh}} | \boldx_m)
%     }} \log p(\boldx_{m'} | \boldz_{\text{sh}}), \text{ if } m \ne m',
%     \end{cases}
% \end{align*}
where ``$\text{sg}$'' refers to the stop-gradient operator. 
Note that as a result, $\zshared$ can still be \emph{used} for same-view predictions, only its \emph{updates} are restricted to cross-modal information. 
% The modified objective remains a valid lower bound on the log marginal likelihood (the stop-gradient can be thought of as a modification of the variational family). 
We refer to this modified objective as MMVAE++. 

\subsection{Supervised multimodal VAEs}

Here, we now focus on the (arguably rare) scenarios where we have access to labelled information $\boldy$ that should reflect the \emph{shared} structure across modalities. For example, in biomedical data, these labels could indicate healthy/disease group, the presence of a particular mutation, or disease subtype. 
Note that our goal here is \emph{not} learning a classifier. Instead, our research question is whether having access to relevant label information can improve the model's ability to identify shared latent structure in the challenging scenarios where standard (fully unsupervised) representation learning fails. %, i.e.\ whether the labels can aid representation learning. 

% utilising these labels for representation learning. 

In order to incorporate label information in the model so that it would encourage $\zshared$ to reflect the truly shared information, we propose to augment the ELBO as follows
\begin{align*}
    \tilde{\calL} := \calL + \beta \, 
    \E_{\boldz_\text{sh} \sim q(\boldz_{\text{sh}} | \boldx_m)} p(\boldy | \boldz)
\end{align*}
where $p(\boldy | \boldz)$  is parameterised by a \emph{linear} mapping, $\calL$ denotes the objective function of any of the previously discussed multimodal VAE, and $\beta$ is a hyperparameter\footnote{
    We note that the choice $\beta=1$ corresponds to the lower bound on the log-marginal $ p(\boldx_{1:M}, \boldy)$, but in practice, we can use $\beta \gg 1$ to upweight the relative importance of the labels, in contrast to other features. Thus, we recommend to choose $\beta$ so that its magnitude is aligned with the magnitude of $\max_m \dim(\boldx_{m})$. In our experiments, we use $\beta = 10^3$. 
}. 
As a result, we can obtain supervised versions of all models which we refer to as sup-MVAE, sup-MoPoE and sup-MMVAE(++).

\section{Results}

\textbf{Goals:}
The goals of our experiments are two-fold. 
First, we want to systematically investigate the inductive biases and capabilities of multimodal VAEs to learn shared latent factors in challenging realistic scenarios, including imbalanced modalities as well as varying levels of modality-specific variation. 
Second, we aim to characterise the failure modes of existing models, i.e.\ the circumstances when they start to break, and whether MMVAE++ will exhibit more desirable behaviour. 
Our experiments focus on tabular data settings and real-world illustrations will use 'omics data sets.

\textbf{Shared and private feature sets:}
In order to distinguish truly shared signal from modality-specific variation, we follow the setup introduced in Figure~\ref{fig:figure1}A, where features are categorised as either ``shared'' or ``private''. 
In synthetic data experiments where we have access to ground truth generative latent factors, it is straightforward to generate such feature sets. 
But in real omics data where ground truth is unknown, we will rely on external labels or biomarkers, which let us create feature sets that would be driven by the same underlying biological phenomenon. As a result, we can partition a modality $\boldx_m$ into two components $\boldx_m = [\boldx_m^{\text{shared}}, \boldx_{m}^{\text{other}}]$. 
In our evaluations, we would like to control the respective dimensionalities. To create feature sets of given sizes $P_m \le \dim(\boldx_m^{\text{shared}})$ and $Q_m \le \dim(\boldx_m^{\text{other}})$, we subsample $P_m$ and $Q_m$ features from the two respective subsets. 

\textbf{Model specification:}
For all versions of multimodal VAEs, we use an identical setup and architecture, and we adopt a structured latent space $\boldz = [\zprivateone, \zshared, \zprivatetwo]$ throughout. We compare MMVAE++ with three baselines: MVAE, MoPoE-VAE, and MMVAE\footnote{To be precise, MVAE and MMVAE with a structured latent space correspond to \citep{lee_private-shared_2021} and \citep{palumbo_mmvae_2022} respectively as explained in Section~\ref{sec:private_and_shared}. Throughout the experiments, we will refer to the terms MVAE, MoPoE-VAE and MMVAE as the respective models with a structured latent space $\boldz = [\zprivateone, \zshared, \zprivatetwo]$.} with varying latent dimensionalities. 
In the synthetic examples, we consider both correctly specified and misspecified latent structure. 
Using notation ``$\dim(\zprivateone) + \dim(\zshared) + \dim(\zprivatetwo)$'', in our real data experiments, we consider dimensionalities ``$2+2+2$'', ``$3+3+3$'', and ``$4+4+4$''. 
Our implementation is available in \url{https://github.com/kasparmartens/shared-private-multimodalVAE}.
% Code for all models and experiments is available in [redacted]. 

\textbf{Evaluation metrics:}
We will evaluate how successfully a method is able to learn shared latent structure by quantifying its cross-view prediction performance. 
In all our examples, we use $M=2$ modalities, all of which contain continuous measurements. 
We quantify the predictive performance by using the point-estimates of $p(\boldx_1 | \boldx_2)$ and $p(\boldx_2 | \boldx_1)$, as outlined in~\eqref{eq:crossmodal_pred}, and report the fraction of variance explained $R^2$ for every feature. 
Note that we would expect high $R^2$ values only for the ``shared'' feature set. By definition, the ``other'' set should not contain any shared signal, and therefore we would expect cross-modal prediction to produce $R^2 \approx 0$. 
In experiments where external labels $\boldy$ are available, we additionally report the area under the ROC curve (AUC) to quantify how linearly separable the classes are in the shared latent space.  

\subsection{Synthetic examples} \label{sec:synthetic}

To compare methods in scenarios where we know the ground truth, we consider data generated
% To compare methods in scenarios where we understand the data generative mechanism, we consider synthetic examples generated 
from a non-linear latent variable model for two modalities $\boldx_1$ and $\boldx_2$. Every feature $j$ in either modality $\boldx_m^{(j)}$ is generated either from shared latent variables $\boldz^{\text{shared}}$ or private ones $\boldz^{\text{pr}_m}$, with latent-to-feature mappings drawn from the Gaussian Process prior $f^{(j)} \sim \GP(0, k(\cdot))$. For shared features, the kernel is defined on $\boldz^{\text{shared}}$, for private features on $\boldz^{\text{pr}_m}$, essentially corresponding to generating data from a multi-view GPLVM \citep{lawrence_probabilistic_2005, damianou_multi-view_2021}.
% That is, we essentially generate data from a particular multi-view Gaussian Process Latent Variable Model (GPLVM) with pre-specified shared and private latent coordinates \citep{lawrence_probabilistic_2005, damianou_multi-view_2021}.

\begin{wrapfigure}{R}{0.5\textwidth}
     \centering
     \includegraphics[width=0.99\linewidth]{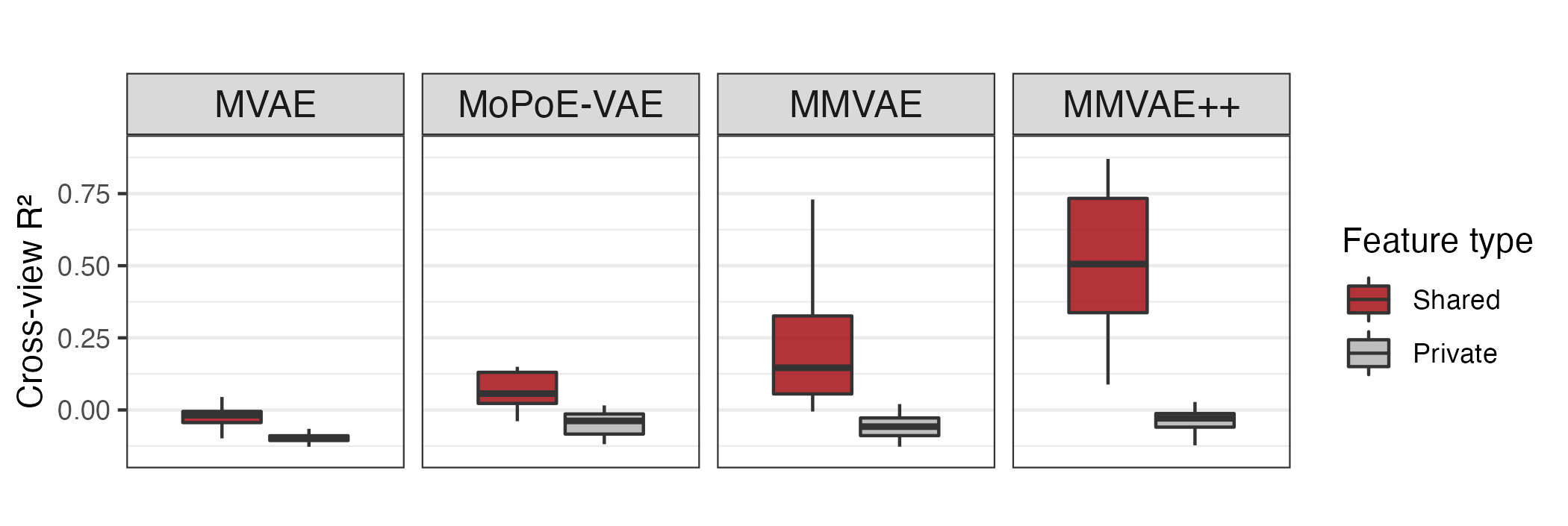}
     \vspace{-2em}
     \caption{
        \textbf{Synthetic GP example:} Cross-view prediction performance ($R^2$), separately for \textit{``shared''} (red) and \textit{``private''} (grey) feature sets. 
        % Boxplots of cross-view prediction accuracy  on synthetic example when the latent dimensionalities in the model are (A) correctly specified, and (B) misspecified. 
     }
     \vspace{-0.5em}
     \label{fig:boxplots_synthetic}
\end{wrapfigure}

In these synthetic examples, we fix the number of \textit{``shared''} features in both modalities $P_1 = P_2 = 10$, and then vary the number of other features $Q_1$ and $Q_2$. We consider two scenarios, generating data from the correctly specified latent structure (``$2+2+2$'') and a misspecified scenario (``$4+2+4$''). 
Figure~\ref{fig:figure1}B shows results for the \textit{``shared''} feature set in the former scenario, where $Q_2$ was fixed to 100 and then we gradually increase $Q_1 \in \{20, 50, 100, \ldots, 1000\}$, and Figure~\ref{fig:boxplots_synthetic} shows a more detailed breakdown (this serves as a sanity check as $R^2 \approx 0$ for \textit{``private''} features is expected behaviour). 
Despite correctly specified latent dimensionalities, the performance of all models except MMVAE++ drops quickly. We also observe that both MoPoE-VAE and MMVAE consistently outperform MVAE, and beyond 50 \textit{``private''} features MMVAE++ is the best performing method. 
In the misspecified case (see Supplementary Figure~\ref{fig:supp_toy_GP}), conclusions remain similar.

\subsection{Chronic Lymphocytic Leukaemia (CLL) study}

\begin{wrapfigure}{R}{0.5\textwidth}
  \vspace{0em}
  \begin{center}
    \includegraphics[width=0.99\linewidth]{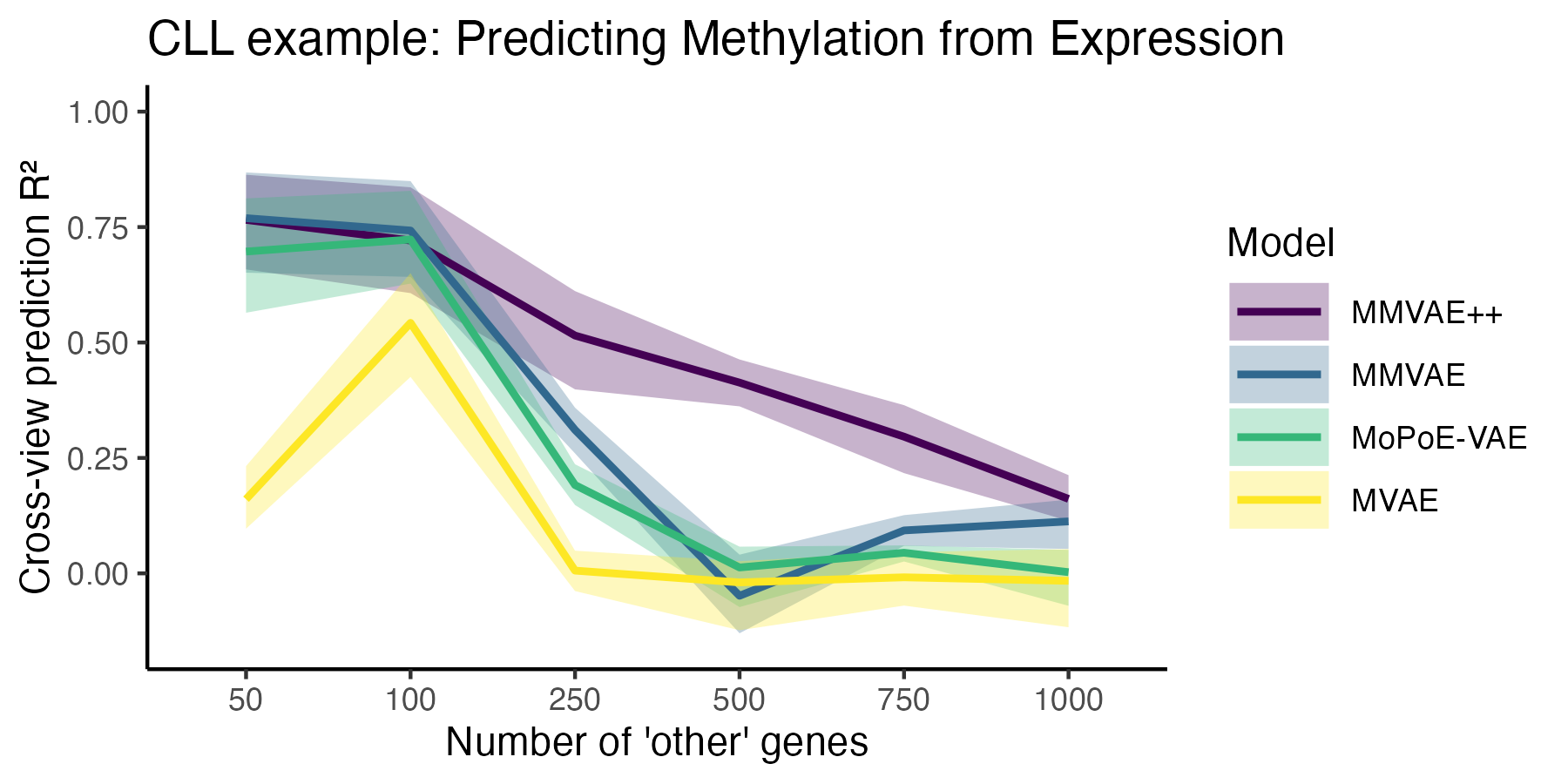}
  \end{center}
  \vspace{-1em}
  \caption{
    \textbf{CLL example:}
        Cross-view $R^2$ for IGHV-related features for varying number of ``other'' genes, when predicting methylation from gene expression. 
        % The shaded area highlights the (5\%, 95\%) quantiles. 
  }
  \vspace{0em}
  \label{fig:CLL_increasing_n_genes}
\end{wrapfigure}

We now consider a study of chronic lymphocytic leukaemia (CLL) \citep{dietrich_drug-perturbation-based_2018, argelaguet_multi-omics_2018}. 
Here, we use gene expression and methylation data
\footnote{We use pre-processed publicly available data from \url{https://github.com/bioFAM/MOFA}}
% subsetting to 
for those $N=126$ patients who have both modalities available. We additionally extract a binary label $\boldy$ which indicates the IGHV mutation status -- a known biomarker for CLL \citep{vasconcelos_gene_2005}. 
We then divide features in both modalities into IGHV-related and the rest\footnote{We do this by ordering all features by their Pearson correlation with IGHV mutation status. To create feature sets of sizes $P_1$ and $Q_1$ respectively, we choose $P_1$ features with the highest absolute correlation and $Q_1$ lowest ones.}. 
We then compare the ability of various multimodal VAEs to find shared latent structure across IGHV-related features (i.e.\ across genes in gene expression data and CpG sites in methylation data). %In gene expression data, the features correspond to genes, whereas in methylation data the features correspond to CpG sites. 

To consider a set of increasingly challenging scenarios, we include $P_1 = P_2 = 10$ IGHV-related features in both modalities, we fix the number of other features in methylation data to $Q_2 = 100$ and then we vary the number of other genes $Q_1 \in \{50, 100, 250, 500, 750, 1000\}$. 
Figure~\ref{fig:CLL_increasing_n_genes} shows the cross-view prediction performance comparison ($R^2$ on held-out subjects) for IGHV-related features when we increase the number of \textit{``other''} genes. 
First, we note that all models exhibit a performance drop. Overall, MVAE is the lowest performing model, confirming the previously discussed inductive biases, while MMVAE++ has highest $R^2$ from 250 \textit{``other''} genes onwards. 
% The above results were obtained using the ``$2+2+2$'' latent space but results with higher dimensional latent spaces have similar conclusions (see Supplementary). 

\begin{table*}[!t]
\begin{center}
\caption{\textbf{AUC for linear classification of IGHV mutation status on the CLL example.} AUC (median $\pm$ sd) for increasing number of ``other'' genes. Evaluated on held-out subjects.
}
\label{tab:CLL_AUC}
\begin{scriptsize}
\begin{sc}
\vskip 0.15in
\begin{tabular}{c|cccc}
    \toprule
    \# other & MVAE & MoPoE-VAE & MMVAE & MMVAE++  \\ 
        \midrule
    50 & \textbf{0.96} \tablehelper{0.01} & \textbf{0.96}  \tablehelper{0.01} & \textbf{0.97} \tablehelper{0.01} & \textbf{0.96} \tablehelper{0.01}  \\
    250 & 0.72 \tablehelper{0.11} & 0.77  \tablehelper{0.07} & 0.76 \tablehelper{0.12} & \textbf{0.83} \tablehelper{0.09}  \\
    500 & 0.45 \tablehelper{0.08} & 0.47  \tablehelper{0.09} & 0.63 \tablehelper{0.16} & \textbf{0.84} \tablehelper{0.21}  \\
    1000 & 0.45 \tablehelper{0.05} & 0.51  \tablehelper{0.05} & 0.43 \tablehelper{0.06} & \textbf{0.62} \tablehelper{0.12}  \\
    \bottomrule 
\end{tabular}
\end{sc}
\end{scriptsize}
\end{center}
\vspace{-2em}
\end{table*}

Table~\ref{tab:CLL_AUC} additionally shows AUC for linear classification of the IGHV mutation status. %, shown for an increasing number of ``other'' genes, separately for all \emph{unsupervised} models (in the left half of the table) and their \emph{supervised} counterparts (on the right-hand side). 
We observe that there is a clear decline in classification performance and learning IGHV-related latent representations becomes increasingly challenging as the number of \textit{``other''} genes increases.  Among unsupervised methods, MMVAE++ remains the best performing at even 500 and 1000 \textit{``other''} genes.  Supervised methods (see Supplementary Table~\ref{tab:supp_CLL_AUC}) also exhibit a decline, albeit more slow (e.g.\ for 1000 \textit{``other''} genes, including labels in MMVAE++ increases the AUC value from 0.62 to 0.76). 
Overall, this example has demonstrated how increasing amounts of modality-specific variation will start to affect the behaviour of all multimodal VAE models, potentially even those that have access to label information. %Among unsupervised approaches, the MMVAE++ stands out in particular as significantly more robust than other unsupervised approaches. 

\subsection{TCGA Breast Cancer study} \label{sec:BRCA}

We now consider the Breast Cancer (BRCA) cohort from The Cancer Genome Atlas \citep{weinstein_cancer_2013}. 
Specifically, we focus on two modalities, gene expression and methylation, restricting our analysis to those $N=479$ patients with both available. 
Breast cancer is a heterogeneous disease: there exist multiple disease subtypes which affect the prognosis as well as response to treatment. 
In our analysis, we focus on two subtypes of breast cancer: oestrogen receptor positive (ER-positive, ER+) and oestrogen receptor negative (ER-negative, ER-). In addition to the two modalities, we also extract the label $\boldy \in \{\text{ER+, ER-}\}$. %These are relatively well understood subtypes: our goal with this analysis is to demonstrate the utility of 

In this experiment, our goal is to test the models' capabilities to extract features (genes and CpGs) that are driven by ER-related biological processes. 
Here, we particularly want to focus on \emph{high-dimensional} and \emph{imbalanced} modalities. 
For this, we include $P_1 = 1000$ ER-related genes and $P_2 = 1000$ ER-related CpGs\footnote{Analogously to the CLL example, we order all features based on the Pearson correlation with ER-status, and then choose $P_1$ with the highest absolute correlation and $Q_1$ lowest ones.}. 
Additionally, we include $Q_1 = 500$ other genes and $Q_2 = \num{25000}$ other CpGs, resulting in total $\dim(\boldx_1) = \num{1500}$ and $\dim(\boldx_2) = \num{26 000}$. %are relatively high-dimensional and imbalanced.  

\begin{figure*}[!ht]
    \centering
    \vspace{-0em}
    \includegraphics[width=\textwidth]{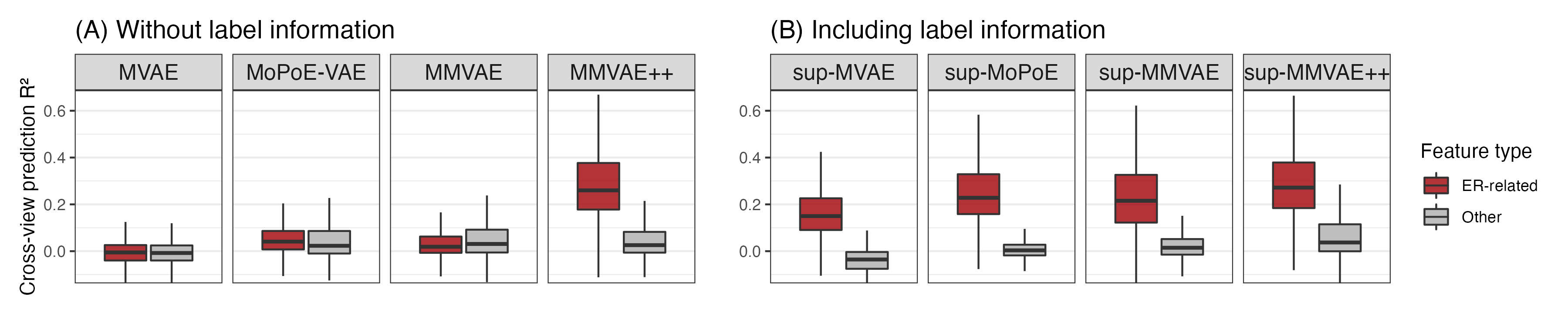}
    \vspace{-1.5em}
    \caption{
        \textbf{BRCA study:} Cross-view prediction accuracy ($R^2$) separately for ER-related (red) and other (grey) feature sets, when predicting expression from methylation. Shown for (A) unsupervised models, and (B) supervised models that have access to the ER-status label.  %The features have been grouped into ER-related (red) and other (grey). We consider (A) unsupervised models, and (B) supervised models that have access to the ER-status label. 
    }
    \vspace{-0.5em}
    \label{fig:BRCA_boxplots}
\end{figure*}

Boxplots in Figure~\ref{fig:BRCA_boxplots} compare the performance of multimodal VAEs in this challenging scenario, with ER-related features being shown in red (see Supp.~Figure~\ref{fig:supp_BRCA_boxplots} for expanded results with higher dimensional latent spaces). 
Among unsupervised approaches, only MMVAE++ has managed to successfully predict many ER-related genes with a median $R^2 = 0.26$ (the second best is MoPoE-VAE with $R^2 = 0.04$). 
% Among unsupervised approaches, both MVAE and MMVAE have a median $R^2 \approx 0$. MVAE++ has a slightly higher median value of 0.04, and only MMVAE++ has managed to successfully predict many ER-related genes with a median $R^2$ value of 0.26. 
This effect is even more clear from Supplementary Figure~\ref{fig:BRCA_featurelevel} where cross-prediction $R^2$ values are shown for selected individual genes. 
For example, for ESR1 -- the gene that encodes the oestrogen receptor -- MMVAE++ achieves a 63 percentage point increase over MMVAE. 
These findings are complemented by AUC values for ER+/ER- label classification in Table~\ref{tab:BRCA_AUC}, where MMVAE++ clearly outperforms other unsupervised approaches. 

\begin{table*}[!h]
\begin{center}
\caption{
\textbf{AUC for linear classification of ER+ and ER- on the BRCA example. }
%Evaluated on held-out data, showing median $\pm$ s.d.
}
\label{tab:BRCA_AUC}
\begin{scriptsize}
\begin{sc}
\begin{tabular}{cccc|cccc}
\toprule
 MVAE & MoPoE-VAE & MMVAE & MMVAE++ & sup-MVAE & sup-MoPoE & sup-MMVAE & sup-MMVAE++ \\ 
 \midrule
 0.62 \plusminus{0.1} & 0.61  \plusminus{0.1} & 0.59 \plusminus{0.1} & \textbf{0.91} \plusminus{0.01} & 0.88\plusminus{0.04} & \textbf{0.92}\plusminus{0.05} & 0.88 \plusminus{0.07} & \textbf{0.93} \plusminus{0.03} \\
\bottomrule
\end{tabular}
\end{sc}
\end{scriptsize}
\end{center}
\vspace{-1.5em}
\end{table*}

When considering supervised methods, we see that incorporating ER-status information has boosted the performance of all models, with the differences between them being small. On the one hand, this suggests that having access to an informative label could have been sufficient to capture ER-related signal in the latent space. But more importantly, it is perhaps surprising that MMVAE++ was able to perform similarly, despite having no access to the label. This can be very useful in scenarios where there are no \textit{a priori} labels on the latent structure that we would want to learn or discover. 

\subsection{Single cell RNA-seq and ATAC-seq experiment}

% So far, we have considered studies involving patient-level (bulk multi-omics) data with moderate sample sizes. In contrast, increasingly popular single cell multi-omic technologies typically produce data sets that are orders of magnitude larger.

So far, we have considered studies involving patient-level (bulk multi-omics) data with moderate sample sizes. In contrast, increasingly popular \emph{single-cell} multi-omic technologies %that let us simultaneously obtain various molecular measurements %(e.g.\ abundance of proteins, or chromatin structure alongside with gene expression) 
% on single cell resolution, 
typically produce orders of magnitude larger data sets.
To demonstrate that our findings equally apply to large data regimes, we now consider a dataset with chromatin accessibility (scATAC-seq) and gene expression (scRNA-seq) measurements across $N=10 032$ peripheral blood mononuclear cells (PBMCs)\footnote{Data and pre-processing from {\tiny \href{https://raw.githack.com/bioFAM/MOFA2_tutorials/master/R_tutorials/10x_scRNA_scATAC.html}{\nolinkurl{raw.githack.com/bioFAM/MOFA2_tutorials/master/R_tutorials/10x_scRNA_scATAC.html}}}}.

It is expected that both modalities capture \emph{cell type} information. Here, we focus on a class of immune cells, \textit{naive CD8 T~cells},  
%, so different cell types should be separable in shared latent space
% Some cell types are relatively easy to distinguish, but here we wanted to focus on a more difficult task, therefore we chose \textit{naive CD8 T~cells}, a class of immune cells, 
with the goal to compare how well different models are able to identify relevant features (from gene expression as well as chromatin peaks) and encode this information in the shared latent space. Analogously to previous experiments, we include $P_1 = P_2 = 10$ features that are most predictive of this cell type, while including $Q_2 = 5000$ other peaks and gradually increasing the number of other genes $Q_1 \in \{10,  100, 1000\}$. 
Table~\ref{tab:cell_types} shows that there is sufficient information in the data to near-perfectly identify the cell type of interest: with 10 other genes, this is the case for all models except MVAE. However, when increasing the number of other features, the capability of both MoPoE-VAE and MMVAE to classify \textit{naive CD8 T cells} starts to decrease, whereas MMVAE++ exhibits only a small decrease in performance. This is comfirmed by shared latent space visualisations in Supplementary Figure~\ref{fig:supp_single_cell}. 
All \emph{supervised} multimodal VAEs achieve near-perfect separation (with AUC 0.99 {\scriptsize $(\pm 0.001)$} in all cases), demonstrating how relevant labels can help shape what is encoded in the shared latent space. %, but also showing that the gap with an unsupervised approach, MMVAE++, remains relatively small. 
% Shared latent space visualisations are shown in Supplementary Figure~\ref{fig:supp_single_cell}. 

\begin{table*}[!h]
\vspace{-1em}
\begin{center}
\caption{\textbf{AUC for linear classification of \textit{naive CD8 T cells}.} AUC (median $\pm$ standard deviation), when increasing the number of ``other'' genes (in the first column) in the gene expression modality.
}
\label{tab:cell_types}
\begin{scriptsize}
\begin{sc}
\vskip 0.05in
\begin{tabular}{c|cccc}
    \toprule
    \# other genes & MVAE & MoPoE-VAE & MMVAE & MMVAE++  \\ 
        \midrule
    % 10 & 0.64 \tablehelper{0.12} & 0.79 \tablehelper{0.01} & 0.86 \tablehelper{0.01} & \textbf{0.96} \tablehelper{0.01} \\
    10 & 0.68 \tablehelper{0.05} & \textbf{0.98} \tablehelper{0.02} & \textbf{0.97} \tablehelper{0.02} & \textbf{0.98} \tablehelper{0.02} \\
    100 & 0.61 \tablehelper{0.13} & 0.82 \tablehelper{0.03} & 0.87 \tablehelper{0.05} & \textbf{0.97} \tablehelper{0.06} \\
    1000 & 0.63 \tablehelper{0.05} & 0.79 \tablehelper{0.10} & 0.81 \tablehelper{0.13} & \textbf{0.91} \tablehelper{0.02} \\
    \bottomrule 
\end{tabular}
\end{sc}
\end{scriptsize}
\end{center}
\vspace{-1em}
\end{table*}

\section{Conclusions}

This work investigated optimal design choices for the implementation of multimodal VAEs with shared and private latent factors. Using multi-omics applications, we have highlighted a problem setting that has turned out to be challenging for existing multimodal VAEs, and discussed how inductive biases in those models relate to their ability to reliably learn shared latent factors in scenarios where modality-specific variation dominates. 
We have proposed a modification MMVAE++ and shown its enhanced ability to learn shared latent structure.  

Multimodal VAEs have been utilised widely for data integration tasks. We believe that our findings highlight the need for careful consideration of the design of these models and their intended use. Our experiments particularly illustrate that when data modalities are imbalanced and/or if the latent structure of interest is not dominant, the inferred latent representations may not be reflective of the true private and shared latent processes and might potentially lead to misleading interpretations. It also places a spotlight on the impact of data filtering and pre-processing steps that commonly occur preceding the construction of analysis-ready datasets. We hope this will inspire further work, spanning both theoretical properties as well as applications, and provide an additional resource for practitioners to make a more informed choice when choosing what type of multimodal VAE to use.

\bibliography{references}

\begin{thebibliography}{38}
\providecommand{\natexlab}[1]{#1}
\providecommand{\url}[1]{\texttt{#1}}
\expandafter\ifx\csname urlstyle\endcsname\relax
  \providecommand{\doi}[1]{doi: #1}\else
  \providecommand{\doi}{doi: \begingroup \urlstyle{rm}\Url}\fi

\bibitem[Stuart and Satija(2019)]{stuart_integrative_2019}
Tim Stuart and Rahul Satija.
\newblock Integrative single-cell analysis.
\newblock \emph{Nature Reviews Genetics}, 20\penalty0 (5):\penalty0 257--272, May 2019.
\newblock ISSN 1471-0064.
\newblock \doi{10.1038/s41576-019-0093-7}.
\newblock URL \url{https://www.nature.com/articles/s41576-019-0093-7}.
\newblock Number: 5 Publisher: Nature Publishing Group.

\bibitem[Hasin et~al.(2017)Hasin, Seldin, and Lusis]{hasin_multi-omics_2017}
Yehudit Hasin, Marcus Seldin, and Aldons Lusis.
\newblock Multi-omics approaches to disease.
\newblock \emph{Genome biology}, 18\penalty0 (1):\penalty0 1--15, 2017.
\newblock Publisher: BioMed Central.

\bibitem[Krassowski et~al.(2020)Krassowski, Das, Sahu, and Misra]{krassowski_state_2020}
Michal Krassowski, Vivek Das, Sangram~K. Sahu, and Biswapriya~B. Misra.
\newblock State of the {Field} in {Multi}-{Omics} {Research}: {From} {Computational} {Needs} to {Data} {Mining} and {Sharing}.
\newblock \emph{Frontiers in Genetics}, 11, 2020.
\newblock ISSN 1664-8021.
\newblock URL \url{https://www.frontiersin.org/articles/10.3389/fgene.2020.610798}.

\bibitem[Hotelling(1936)]{hotelling_relations_1936}
Harold Hotelling.
\newblock Relations between two sets of variates.
\newblock \emph{Biometrika}, 28\penalty0 (3/4):\penalty0 321, 1936.
\newblock Publisher: JSTOR.

\bibitem[Tucker(1958)]{tucker_inter-battery_1958}
Ledyard~R. Tucker.
\newblock An inter-battery method of factor analysis.
\newblock \emph{Psychometrika}, 23\penalty0 (2):\penalty0 111--136, 1958.
\newblock Publisher: Springer.

\bibitem[Klami and Kaski(2006)]{klami_generative_2006}
Arto Klami and Samuel Kaski.
\newblock Generative models that discover dependencies between data sets.
\newblock In \emph{2006 16th {IEEE} {Signal} {Processing} {Society} {Workshop} on {Machine} {Learning} for {Signal} {Processing}}, pages 123--128. IEEE, 2006.

\bibitem[Ek and Lawrence(2009)]{ek_shared_2009}
Carl~Henrik Ek and PHTND Lawrence.
\newblock \emph{Shared {Gaussian} process latent variable models}.
\newblock {PhD} {Thesis}, Citeseer, 2009.

\bibitem[Virtanen et~al.(2012)Virtanen, Klami, Khan, and Kaski]{virtanen_bayesian_2012}
Seppo Virtanen, Arto Klami, Suleiman Khan, and Samuel Kaski.
\newblock Bayesian group factor analysis.
\newblock In \emph{Artificial {Intelligence} and {Statistics}}, pages 1269--1277. PMLR, 2012.

\bibitem[Klami et~al.(2013)Klami, Virtanen, and Kaski]{klami_bayesian_2013}
Arto Klami, Seppo Virtanen, and Samuel Kaski.
\newblock Bayesian {Canonical} correlation analysis.
\newblock \emph{Journal of Machine Learning Research}, 14\penalty0 (4), 2013.

\bibitem[Wang et~al.(2016)Wang, Yan, Lee, and Livescu]{wang_deep_2016}
Weiran Wang, Xinchen Yan, Honglak Lee, and Karen Livescu.
\newblock Deep variational canonical correlation analysis.
\newblock \emph{arXiv preprint arXiv:1610.03454}, 2016.

\bibitem[Gundersen et~al.(2020)Gundersen, Dumitrascu, Ash, and Engelhardt]{gundersen_end--end_2020}
Gregory Gundersen, Bianca Dumitrascu, Jordan~T. Ash, and Barbara~E. Engelhardt.
\newblock End-to-end {Training} of {Deep} {Probabilistic} {CCA} on {Paired} {Biomedical} {Observations}.
\newblock In \emph{Proceedings of {The} 35th {Uncertainty} in {Artificial} {Intelligence} {Conference}}, pages 945--955. PMLR, August 2020.
\newblock URL \url{https://proceedings.mlr.press/v115/gundersen20a.html}.
\newblock ISSN: 2640-3498.

\bibitem[Damianou et~al.(2021)Damianou, Lawrence, and Ek]{damianou_multi-view_2021}
Andreas~C. Damianou, Neil~D. Lawrence, and Carl~Henrik Ek.
\newblock Multi-view {Learning} as a {Nonparametric} {Nonlinear} {Inter}-{Battery} {Factor} {Analysis}.
\newblock \emph{J. Mach. Learn. Res.}, 22\penalty0 (86):\penalty0 1--51, 2021.

\bibitem[Kingma and Welling(2014)]{kingma_auto-encoding_2014}
Diederik~P. Kingma and Max Welling.
\newblock Auto-encoding variational bayes.
\newblock \emph{Proceedings of the International Conference on Learning Representations (ICLR)}, 2014.

\bibitem[Rezende et~al.(2014)Rezende, Mohamed, and Wierstra]{rezende_stochastic_2014}
Danilo~Jimenez Rezende, Shakir Mohamed, and Daan Wierstra.
\newblock Stochastic backpropagation and approximate inference in deep generative models.
\newblock \emph{arXiv preprint arXiv:1401.4082}, 2014.

\bibitem[Lopez et~al.(2018)Lopez, Regier, Cole, Jordan, and Yosef]{lopez_deep_2018}
Romain Lopez, Jeffrey Regier, Michael~B. Cole, Michael~I. Jordan, and Nir Yosef.
\newblock Deep generative modeling for single-cell transcriptomics.
\newblock \emph{Nature methods}, 15\penalty0 (12):\penalty0 1053, 2018.

\bibitem[Lotfollahi et~al.(2023)Lotfollahi, Klimovskaia~Susmelj, De~Donno, Hetzel, Ji, Ibarra, Srivatsan, Naghipourfar, Daza, Martin, Shendure, McFaline‐Figueroa, Boyeau, Wolf, Yakubova, Günnemann, Trapnell, Lopez‐Paz, and Theis]{lotfollahi_predicting_2023}
Mohammad Lotfollahi, Anna Klimovskaia~Susmelj, Carlo De~Donno, Leon Hetzel, Yuge Ji, Ignacio~L Ibarra, Sanjay~R Srivatsan, Mohsen Naghipourfar, Riza~M Daza, Beth Martin, Jay Shendure, Jose~L McFaline‐Figueroa, Pierre Boyeau, F~Alexander Wolf, Nafissa Yakubova, Stephan Günnemann, Cole Trapnell, David Lopez‐Paz, and Fabian~J Theis.
\newblock Predicting cellular responses to complex perturbations in high‐throughput screens.
\newblock \emph{Molecular Systems Biology}, 19\penalty0 (6):\penalty0 e11517, June 2023.
\newblock ISSN 1744-4292.
\newblock \doi{10.15252/msb.202211517}.
\newblock URL \url{https://www.embopress.org/doi/full/10.15252/msb.202211517}.
\newblock Publisher: John Wiley \& Sons, Ltd.

\bibitem[Weinberger et~al.(2022)Weinberger, Lopez, Huetter, and Regev]{weinberger_disentangling_2022}
Ethan Weinberger, Romain Lopez, Jan-Christian Huetter, and Aviv Regev.
\newblock Disentangling shared and group-specific variations in single-cell transcriptomics data with {multiGroupVI}.
\newblock In \emph{Proceedings of the 17th {Machine} {Learning} in {Computational} {Biology} meeting}, pages 16--32. PMLR, December 2022.
\newblock URL \url{https://proceedings.mlr.press/v200/weinberger22a.html}.
\newblock ISSN: 2640-3498.

\bibitem[Minoura et~al.(2021)Minoura, Abe, Nam, Nishikawa, and Shimamura]{minoura_mixture--experts_2021}
Kodai Minoura, Ko~Abe, Hyunha Nam, Hiroyoshi Nishikawa, and Teppei Shimamura.
\newblock A mixture-of-experts deep generative model for integrated analysis of single-cell multiomics data.
\newblock \emph{Cell reports methods}, 1\penalty0 (5):\penalty0 100071, 2021.
\newblock Publisher: Elsevier.

\bibitem[Gayoso et~al.(2021)Gayoso, Steier, Lopez, Regier, Nazor, Streets, and Yosef]{gayoso_joint_2021}
Adam Gayoso, Zoë Steier, Romain Lopez, Jeffrey Regier, Kristopher~L. Nazor, Aaron Streets, and Nir Yosef.
\newblock Joint probabilistic modeling of single-cell multi-omic data with {totalVI}.
\newblock \emph{Nature methods}, 18\penalty0 (3):\penalty0 272--282, 2021.
\newblock Publisher: Nature Publishing Group.

\bibitem[Lotfollahi et~al.(2022)Lotfollahi, Litinetskaya, and Theis]{lotfollahi_multigrate_2022}
Mohammad Lotfollahi, Anastasia Litinetskaya, and Fabian~J. Theis.
\newblock Multigrate: single-cell multi-omic data integration.
\newblock \emph{bioRxiv}, 2022.
\newblock Publisher: Cold Spring Harbor Laboratory.

\bibitem[Cao et~al.(2022)Cao, Fu, Wu, Peng, Nie, Zhang, and Xie]{cao_integrated_2022}
Yingxin Cao, Laiyi Fu, Jie Wu, Qinke Peng, Qing Nie, Jing Zhang, and Xiaohui Xie.
\newblock Integrated analysis of multimodal single-cell data with structural similarity.
\newblock \emph{Nucleic Acids Research}, 50\penalty0 (21):\penalty0 e121--e121, 2022.
\newblock Publisher: Oxford University Press.

\bibitem[Liu et~al.(2022)Liu, Greenberg, and Shomorony]{liu_cvqvae_2022}
Tianyu Liu, Grant Greenberg, and Ilan Shomorony.
\newblock {CVQVAE}: {A} representation learning based method for multi-omics single cell data integration.
\newblock In \emph{Proceedings of the 17th {Machine} {Learning} in {Computational} {Biology} meeting}, pages 1--15. PMLR, December 2022.
\newblock URL \url{https://proceedings.mlr.press/v200/liu22a.html}.
\newblock ISSN: 2640-3498.

\bibitem[Wu and Goodman(2018)]{wu_multimodal_2018}
Mike Wu and Noah Goodman.
\newblock Multimodal generative models for scalable weakly-supervised learning.
\newblock \emph{Advances in Neural Information Processing Systems}, 31, 2018.

\bibitem[Shi et~al.(2019)Shi, Paige, and Torr]{shi_variational_2019}
Yuge Shi, Brooks Paige, and Philip Torr.
\newblock Variational mixture-of-experts autoencoders for multi-modal deep generative models.
\newblock \emph{Advances in Neural Information Processing Systems}, 32, 2019.

\bibitem[Lee and Schaar(2021)]{lee_variational_2021}
Changhee Lee and Mihaela van~der Schaar.
\newblock A {Variational} {Information} {Bottleneck} {Approach} to {Multi}-{Omics} {Data} {Integration}.
\newblock In Arindam Banerjee and Kenji Fukumizu, editors, \emph{The 24th {International} {Conference} on {Artificial} {Intelligence} and {Statistics}, {AISTATS} 2021, {April} 13-15, 2021, {Virtual} {Event}}, volume 130 of \emph{Proceedings of {Machine} {Learning} {Research}}, pages 1513--1521. PMLR, 2021.
\newblock URL \url{http://proceedings.mlr.press/v130/lee21a.html}.

\bibitem[Sutter et~al.(2021)Sutter, Daunhawer, and Vogt]{sutter_generalized_2021}
Thomas~M. Sutter, Imant Daunhawer, and Julia~E. Vogt.
\newblock Generalized {Multimodal} {ELBO}.
\newblock In \emph{International {Conference} on {Learning} {Representations}}, January 2021.
\newblock URL \url{https://openreview.net/forum?id=5Y21V0RDBV}.

\bibitem[Lee and Pavlovic(2021)]{lee_private-shared_2021}
Mihee Lee and Vladimir Pavlovic.
\newblock Private-shared disentangled multimodal vae for learning of latent representations.
\newblock In \emph{Proceedings of the {IEEE}/{CVF} {Conference} on {Computer} {Vision} and {Pattern} {Recognition}}, pages 1692--1700, 2021.

\bibitem[Palumbo et~al.(2022)Palumbo, Daunhawer, and Vogt]{palumbo_mmvae_2022}
Emanuele Palumbo, Imant Daunhawer, and Julia~E. Vogt.
\newblock {MMVAE}+: {Enhancing} the {Generative} {Quality} of {Multimodal} {VAEs} without {Compromises}.
\newblock In \emph{{ICLR} {Workshop} on {Deep} {Generative} {Models} for {Highly} {Structured} {Data}}, 2022.

\bibitem[Argelaguet et~al.(2018)Argelaguet, Velten, Arnol, Dietrich, Zenz, Marioni, Buettner, Huber, and Stegle]{argelaguet_multi-omics_2018}
Ricard Argelaguet, Britta Velten, Damien Arnol, Sascha Dietrich, Thorsten Zenz, John~C. Marioni, Florian Buettner, Wolfgang Huber, and Oliver Stegle.
\newblock Multi-{Omics} {Factor} {Analysis}—a framework for unsupervised integration of multi-omics data sets.
\newblock \emph{Molecular systems biology}, 14\penalty0 (6):\penalty0 e8124, 2018.

\bibitem[Ash et~al.(2021)Ash, Darnell, Munro, and Engelhardt]{ash_joint_2021}
Jordan~T. Ash, Gregory Darnell, Daniel Munro, and Barbara~E. Engelhardt.
\newblock Joint analysis of expression levels and histological images identifies genes associated with tissue morphology.
\newblock \emph{Nature communications}, 12\penalty0 (1):\penalty0 1--12, 2021.
\newblock Publisher: Nature Publishing Group.

\bibitem[Blei et~al.(2017)Blei, Kucukelbir, and McAuliffe]{blei_variational_2017}
David~M. Blei, Alp Kucukelbir, and Jon~D. McAuliffe.
\newblock Variational inference: {A} review for statisticians.
\newblock \emph{Journal of the American Statistical Association}, 112\penalty0 (518):\penalty0 859--877, 2017.

\bibitem[Suzuki et~al.(2016)Suzuki, Nakayama, and Matsuo]{suzuki_joint_2016}
Masahiro Suzuki, Kotaro Nakayama, and Yutaka Matsuo.
\newblock Joint multimodal learning with deep generative models.
\newblock \emph{arXiv preprint arXiv:1611.01891}, 2016.

\bibitem[Hinton(2002)]{hinton_training_2002}
Geoffrey~E. Hinton.
\newblock Training products of experts by minimizing contrastive divergence.
\newblock \emph{Neural computation}, 14\penalty0 (8):\penalty0 1771--1800, 2002.
\newblock Publisher: MIT Press.

\bibitem[Jacobs et~al.(1991)Jacobs, Jordan, Nowlan, and Hinton]{jacobs_adaptive_1991}
Robert~A. Jacobs, Michael~I. Jordan, Steven~J. Nowlan, and Geoffrey~E. Hinton.
\newblock Adaptive mixtures of local experts.
\newblock \emph{Neural computation}, 3\penalty0 (1):\penalty0 79--87, 1991.
\newblock Publisher: MIT Press.

\bibitem[Lawrence(2005)]{lawrence_probabilistic_2005}
Neil Lawrence.
\newblock Probabilistic non-linear principal component analysis with {Gaussian} process latent variable models.
\newblock \emph{Journal of machine learning research}, 6\penalty0 (Nov):\penalty0 1783--1816, 2005.

\bibitem[Dietrich et~al.(2018)Dietrich, Oleś, Lu, Sellner, Anders, Velten, Wu, Huellein, da~Silva~Liberio, and Walther]{dietrich_drug-perturbation-based_2018}
Sascha Dietrich, Ma{\textbackslash}lgorzata Oleś, Junyan Lu, Leopold Sellner, Simon Anders, Britta Velten, Bian Wu, Jennifer Huellein, Michelle da~Silva~Liberio, and Tatjana Walther.
\newblock Drug-perturbation-based stratification of blood cancer.
\newblock \emph{The Journal of clinical investigation}, 128\penalty0 (1):\penalty0 427--445, 2018.
\newblock Publisher: Am Soc Clin Investig.

\bibitem[Vasconcelos et~al.(2005)Vasconcelos, De~Vos, Vallat, Rème, Lalanne, Wanherdrick, Michel, Nguyen-Khac, Oppezzo, Magnac, Maloum, Ajchenbaum-Cymbalista, Troussard, Leporrier, Klein, Dighiero, and Davi]{vasconcelos_gene_2005}
Y.~Vasconcelos, J.~De~Vos, L.~Vallat, T.~Rème, A.~I. Lalanne, K.~Wanherdrick, A.~Michel, F.~Nguyen-Khac, P.~Oppezzo, C.~Magnac, K.~Maloum, F.~Ajchenbaum-Cymbalista, X.~Troussard, M.~Leporrier, B.~Klein, G.~Dighiero, and F.~Davi.
\newblock Gene expression profiling of chronic lymphocytic leukemia can discriminate cases with stable disease and mutated {Ig} genes from those with progressive disease and unmutated {Ig} genes.
\newblock \emph{Leukemia}, 19\penalty0 (11):\penalty0 2002--2005, November 2005.
\newblock ISSN 1476-5551.
\newblock \doi{10.1038/sj.leu.2403865}.
\newblock URL \url{https://www.nature.com/articles/2403865}.
\newblock Number: 11 Publisher: Nature Publishing Group.

\bibitem[Weinstein et~al.(2013)Weinstein, Collisson, Mills, Shaw, Ozenberger, Ellrott, Shmulevich, Sander, Stuart, and Network]{weinstein_cancer_2013}
John~N. Weinstein, Eric~A. Collisson, Gordon~B. Mills, Kenna R.~Mills Shaw, Brad~A. Ozenberger, Kyle Ellrott, Ilya Shmulevich, Chris Sander, Joshua~M. Stuart, and Cancer Genome Atlas~Research Network.
\newblock The cancer genome atlas pan-cancer analysis project.
\newblock \emph{Nature genetics}, 45\penalty0 (10):\penalty0 1113, 2013.

\end{thebibliography}
% \bibliographystyle

%%%%%%%%%%%%%%%%%%%%%%%%%%%%%%%%%%%%%%%%%%%%%%%%%%%%%%%%%%%%%%%%%%%%%%%%%%%%%%%
%%%%%%%%%%%%%%%%%%%%%%%%%%%%%%%%%%%%%%%%%%%%%%%%%%%%%%%%%%%%%%%%%%%%%%%%%%%%%%%
% APPENDIX
%%%%%%%%%%%%%%%%%%%%%%%%%%%%%%%%%%%%%%%%%%%%%%%%%%%%%%%%%%%%%%%%%%%%%%%%%%%%%%%
%%%%%%%%%%%%%%%%%%%%%%%%%%%%%%%%%%%%%%%%%%%%%%%%%%%%%%%%%%%%%%%%%%%%%%%%%%%%%%%
\newpage
\appendix
\onecolumn

\renewcommand{\thesection}{S\arabic{section}}
\renewcommand{\thefigure}{S\arabic{figure}}
\renewcommand{\theHfigure}{S\arabic{figure}}
\renewcommand{\thetable}{S\arabic{table}}
\renewcommand{\theHtable}{S\arabic{table}}

\setcounter{figure}{0}
\setcounter{table}{0}

% \counterwithin{figure}{section}
% \counterwithin{table}{section}

\section*{Supplementary Material}

\section{Partitioned latent space and cross-modal prediction}

 \begin{figure}[!h]
     \centering
     \includegraphics[width=0.6\columnwidth]{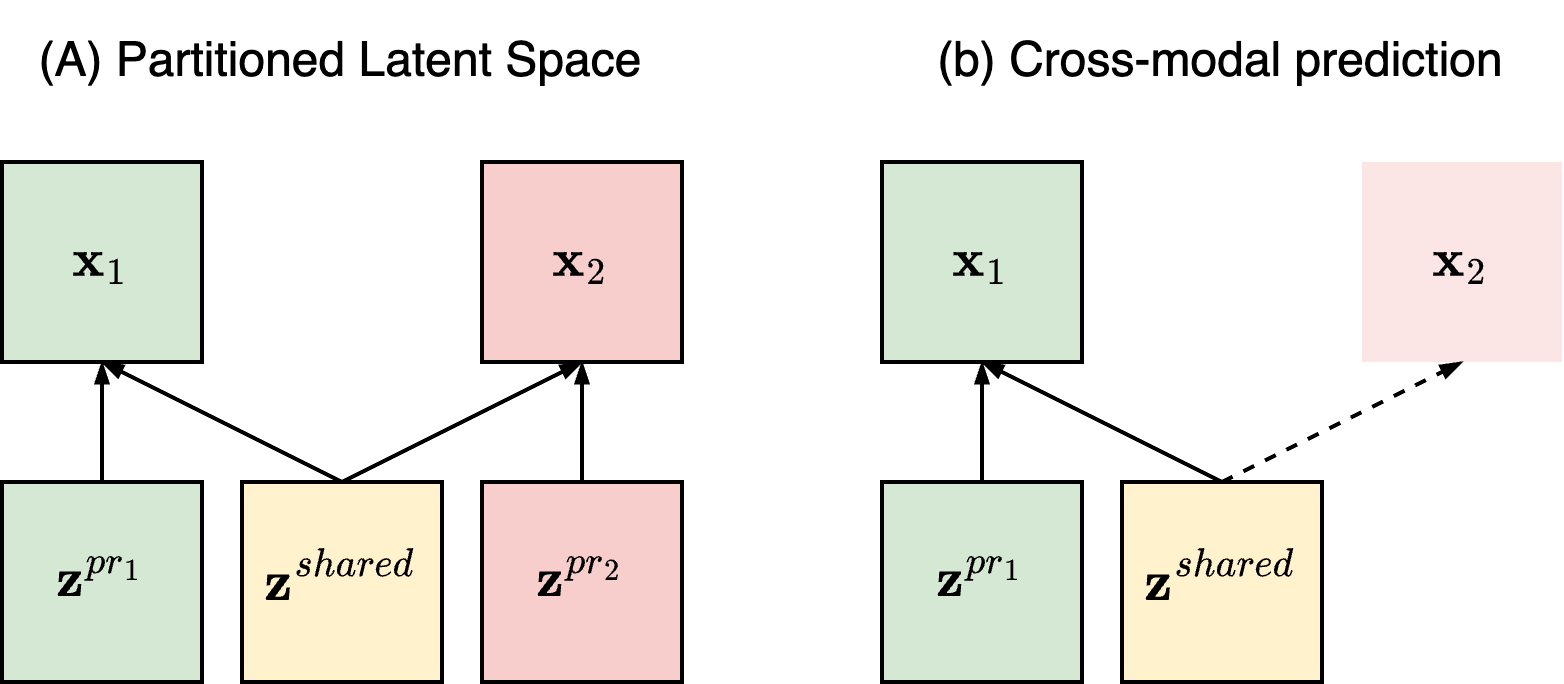}
     \caption{
         Illustration of a partitioned latent space $\boldz = [\zprivateone, \zshared, \zprivatetwo]$ in a multimodal VAE. 
         The use of (a) shared and private latent variables allows for (b) cross-modal prediction via the shared component when one modality maybe missing at train or test time. 
     }
     \label{fig:crossmodal_prediction}
 \end{figure}

\section{Synthetic GP data examples: additional results}

\begin{figure}[!h]
    \centering
    \includegraphics[width=\textwidth]{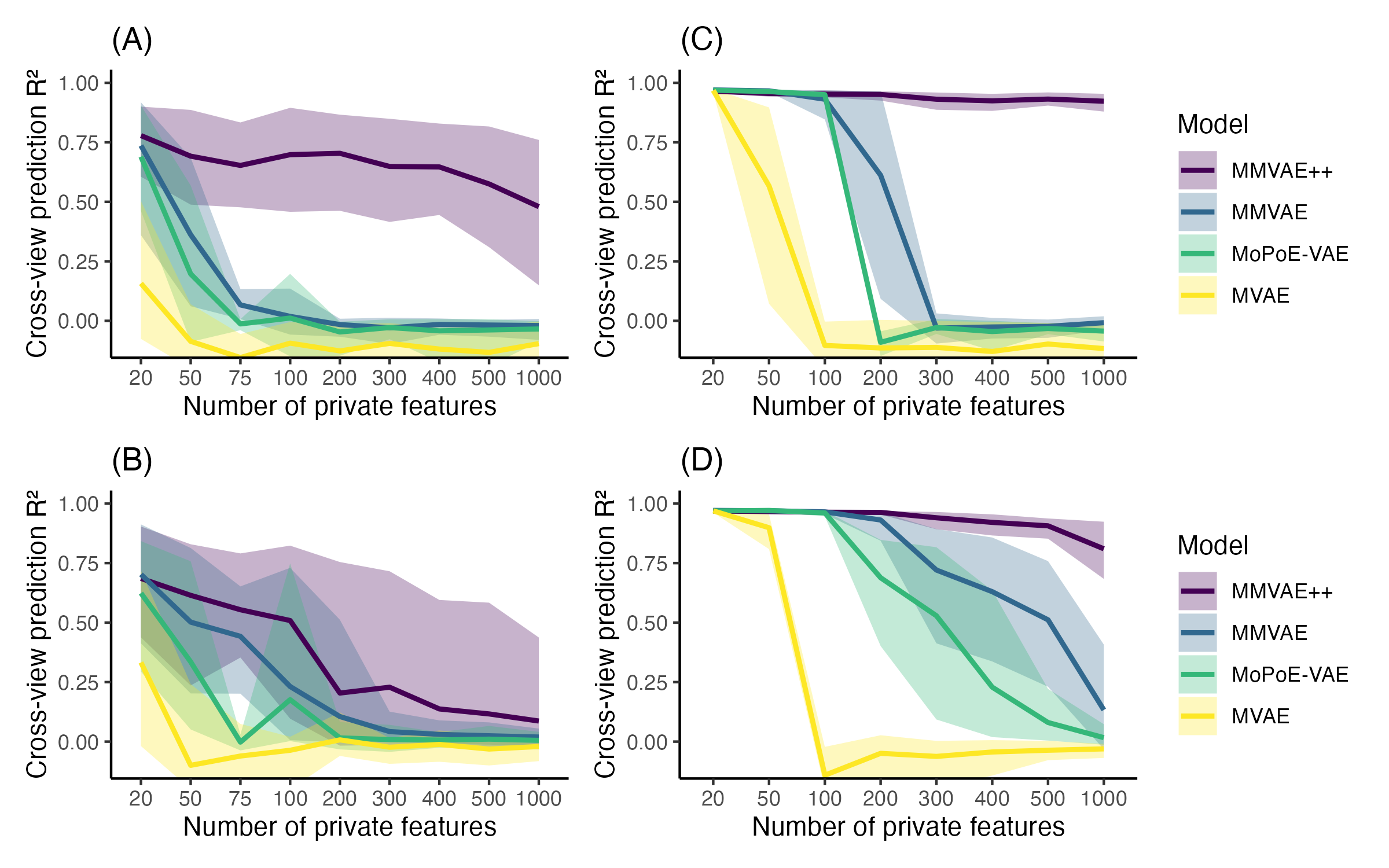}
    \caption{Full set of results for the synthetic GP data experiment, covering (A-B) the correctly specified and (C-D) misspecified latent dimensionalities. Cross-modal prediction performance ($R^2$) shown for (A, C) predicting the first modality from the second one, and (B, D) vice versa.}
    \label{fig:supp_toy_GP}
\end{figure}

\newpage
\section{BRCA example: additional results}

\begin{figure}[!h]
    \centering
    \includegraphics[width=0.6\columnwidth]{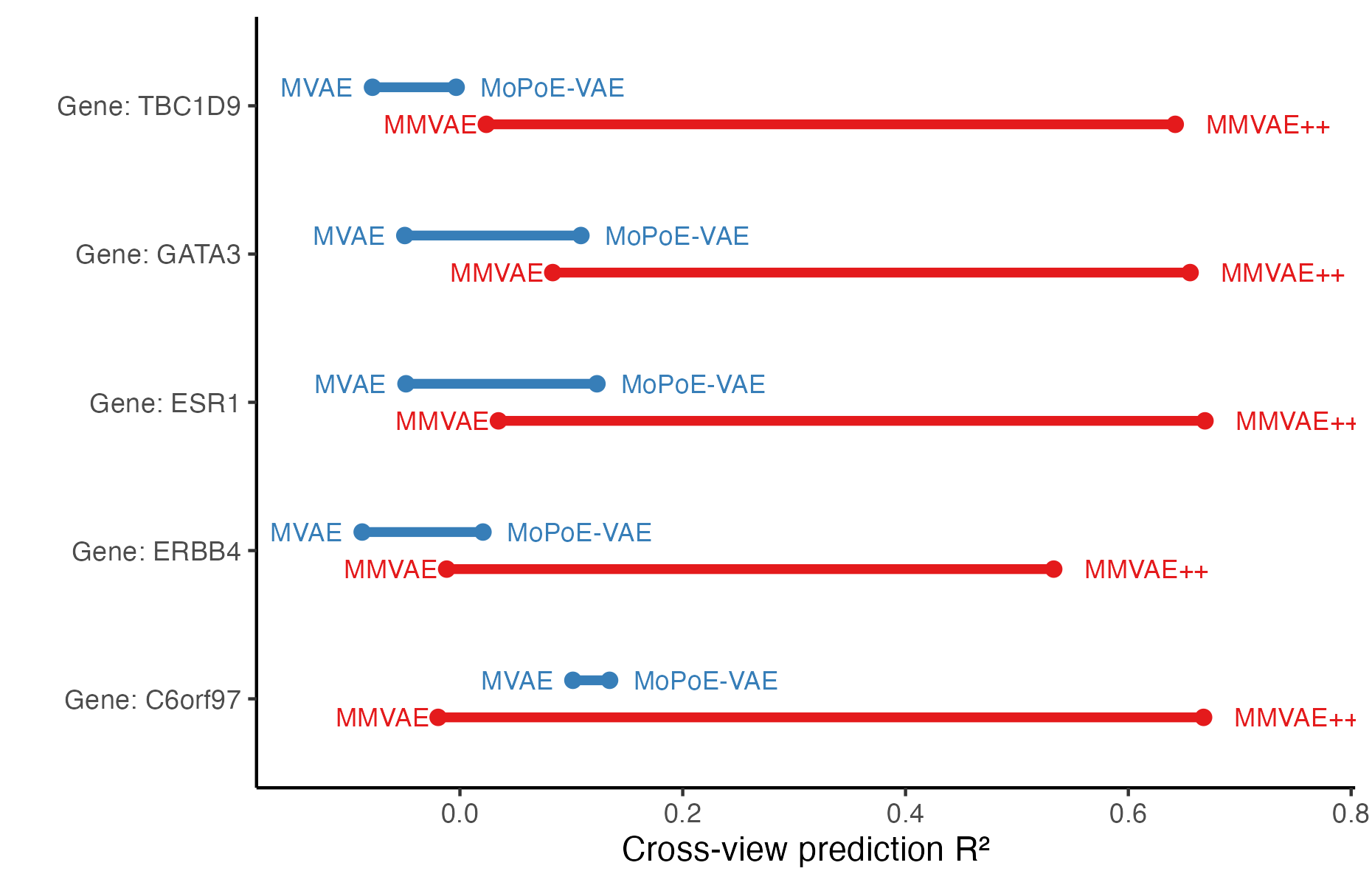}
    \caption{
        \textbf{Cross-modal prediction ($R^2$) for top ER-related genes.} MoPoE-VAE consistently improves upon MVAE, and our proposal MMVAE++ significantly outperforms MMVAE  (blue and red lines highlight the respective gaps). 
        For example, for ESR1, a gene that is known to encode oestrogen receptor, MMVAE++ achieves a 63 percentage point increase over MMVAE. 
    }
    \label{fig:BRCA_featurelevel}
\end{figure}

\begin{figure}[!h]
     \centering
     \includegraphics[width=\textwidth]{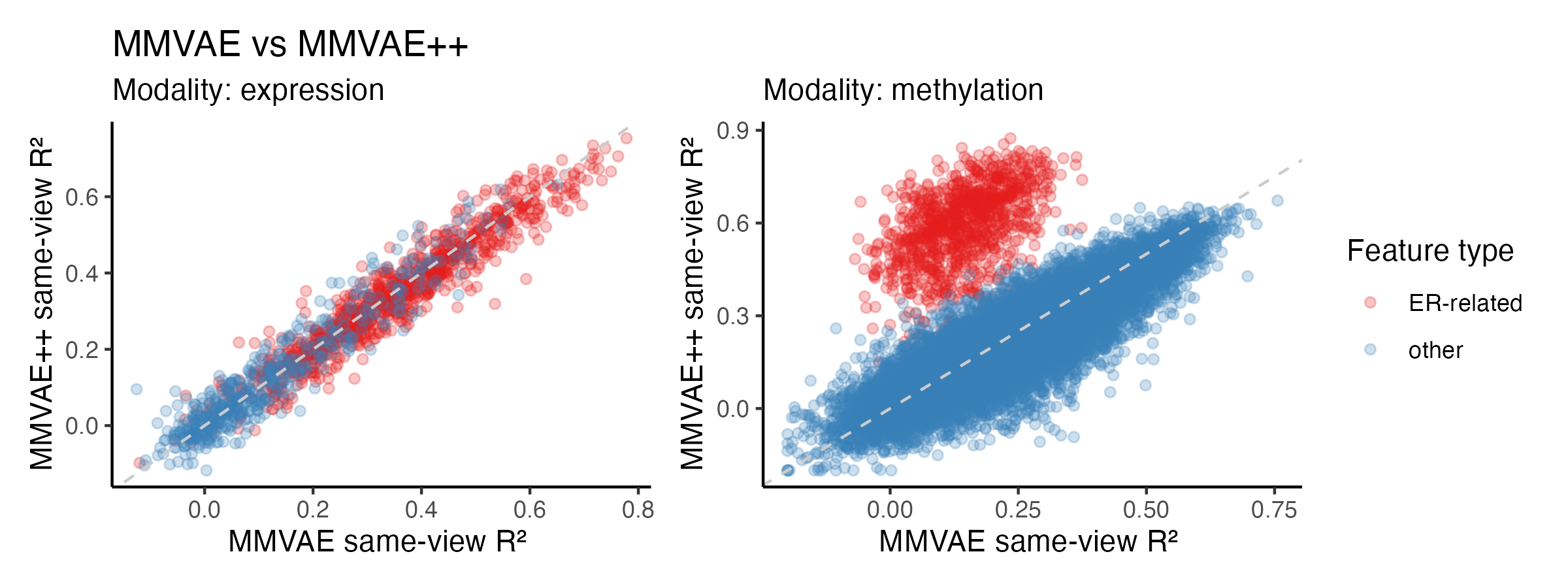}
     \caption{
     \textbf{The improvements in cross-modal prediction do not compromise same-view prediction quality.} While most of our analysis focuses on the quality of \emph{shared} latent factors, we would also want to be sure that the improvements that MMVAE++ brings do not compromise \emph{private} latent factor quality. In real-life examples, this cannot be measured directly, but here we do it indirectly via \emph{same-view} prediction. That is, we obtain predictions for modality $\boldx_m$ under the encoding distribution $q(\boldz | \boldx_m)$. 
     If the models performed similarly, we would expect the values to be symmetric w.r.t. the diagonal line. This is indeed the case for the first modality (expression). But notably, for the second modality, our modifications have improved same-view prediction on the ER-related feature set, while not exhibiting a decrease on the ``other'' feature set.
     }
     \label{fig:supp_BRCA_same_view}
 \end{figure}

\begin{table*}[!h]
\begin{center}
\caption{ \textbf{AUC for linear classification (ER+ and ER-) on the BRCA example}, now expanded across varying latent dimensionalities, i.e.\ $[\text{dim}(\zprivateone), \text{dim}(\zshared),  \text{dim}(\zprivatetwo)]$ values in $[2, 2, 2], [3, 3, 3]$ and $[4, 4, 4]$. 
Evaluated on held-out data, showing median $\pm$ s.d.}
\label{tab:supp_BRCA_AUC}
\begin{scriptsize}
\begin{sc}
\vskip 0.15in
\begin{tabular}{ccccc|cccc}
\toprule
& MVAE & MoPoE-VAE & MMVAE & MMVAE++ & sup-MVAE & sup-MoPoE & sup-MMVAE & sup-MMVAE++  \\ 
 \midrule
``2+2+2'' & 0.62 \plusminus{0.11} & 0.61  \plusminus{0.13} & 0.59 \plusminus{0.08} & \textbf{0.91} \plusminus{0.01}  & 0.88\plusminus{0.04} & \textbf{0.92}\plusminus{0.05} & 0.88 \plusminus{0.07} & \textbf{0.93} \plusminus{0.03}  \\
``3+3+3'' & 0.61 \plusminus{0.08} & 0.88  \plusminus{0.07} & 0.85 \plusminus{0.02} & \textbf{0.94} \plusminus{0.03}  & 0.89 \plusminus{0.04} & \textbf{0.91}\plusminus{0.03} & \textbf{0.94} \plusminus{0.03} & \textbf{0.91} \plusminus{0.03}  \\
``4+4+4'' & 0.87 \plusminus{0.03} & 0.90  \plusminus{0.08} & \textbf{0.92} \plusminus{0.03} & \textbf{0.94} \plusminus{0.03}  & \textbf{0.91} \plusminus{0.08} & \textbf{0.91}\plusminus{0.03} & \textbf{0.94} \plusminus{0.03} & \textbf{0.93} \plusminus{0.03}  \\
\bottomrule
\end{tabular}
\end{sc}
\end{scriptsize}
\end{center}
\vskip -0.1in
\end{table*}

 \begin{figure}[!ht]
     \centering
     \includegraphics[width=\columnwidth]{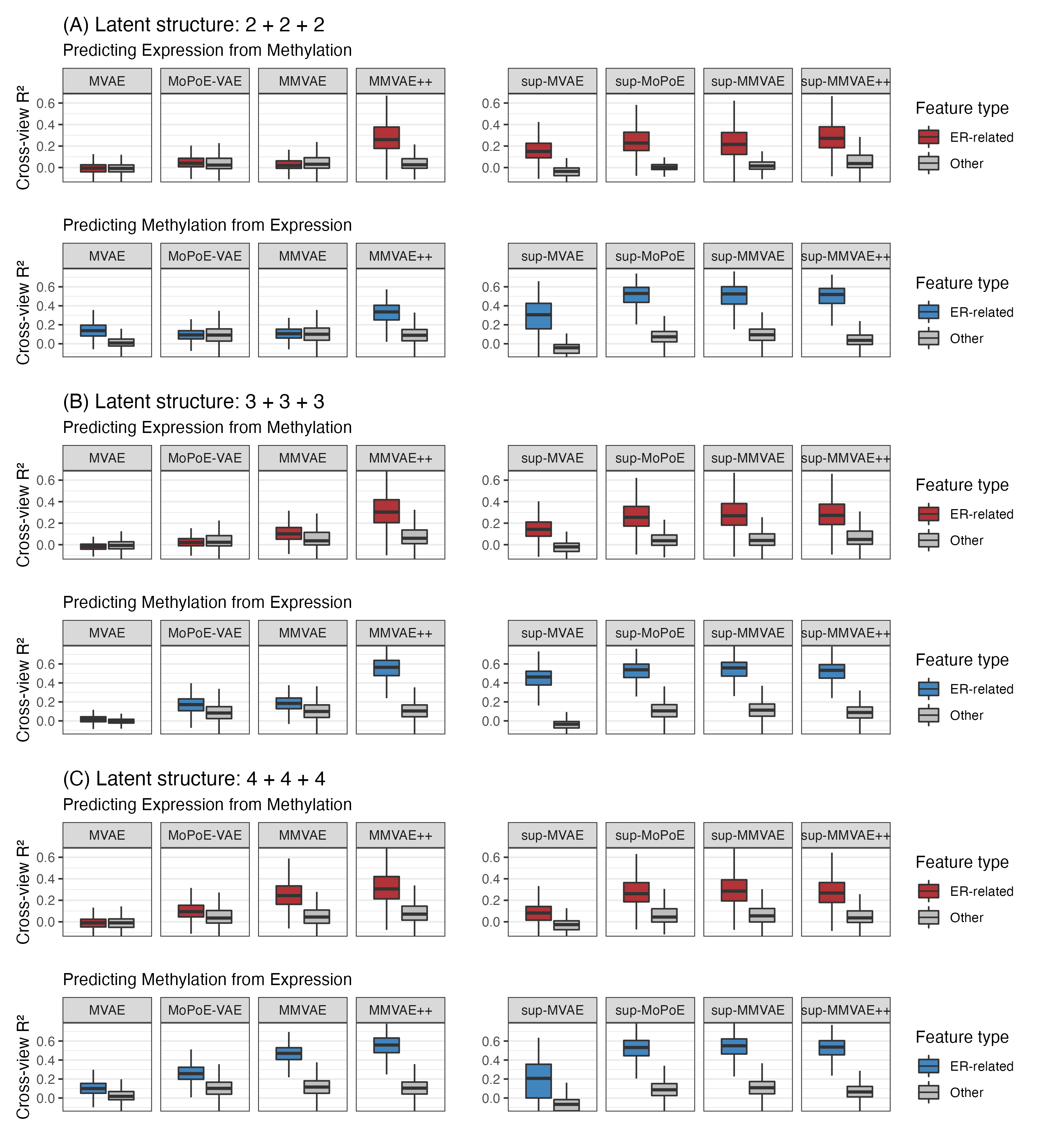}
     \caption{
       BRCA study expanded, now including results for both cross-modal prediction tasks across a range of latent dimensionalities, varying $[\text{dim}(\zprivateone), \text{dim}(\zshared),  \text{dim}(\zprivatetwo)]$ values in $[2, 2, 2], [3, 3, 3]$ and $[4, 4, 4]$. Results for predicting expression from methylation are shown in red, and predicting methylation from expression are shown in blue. 
     }
     \label{fig:supp_BRCA_boxplots}
 \end{figure}

\newpage
\clearpage

\section{CLL example: additional results}

\begin{table*}[!h]
\begin{center}
\caption{\textbf{AUC for linear classification of IGHV mutation status on the CLL example: full set of results.} AUC (median $\pm$ standard deviation), when increasing the number of ``other'' genes (in the first column) in the gene expression modality. Evaluated on held-out subjects.
}
\label{tab:supp_CLL_AUC}
\begin{scriptsize}
\begin{sc}
\vskip 0.15in
% \begin{tabular}{c|cccc|cccc}
%     \toprule
%     \# other & MMVAE & MMVAE++ & MVAE & MVAE++ & sup-MMVAE & sup-MMVAE++ & sup-MVAE & sup-MVAE++ \\ 
%      \midrule
%     50 & \textbf{0.97} \tablehelper{0.01} & \textbf{0.96} \tablehelper{0.01} & \textbf{0.96} \tablehelper{0.01} & \textbf{0.96}  \tablehelper{0.01} & \textbf{0.93} \tablehelper{0.04} & \textbf{0.93} \tablehelper{0.04} & \textbf{0.93} \tablehelper{0.04} & \textbf{0.93}\tablehelper{0.04} \\
%     250 & 0.76 \tablehelper{0.12} & \textbf{0.83} \tablehelper{0.09} & 0.72 \tablehelper{0.11} & 0.77  \tablehelper{0.07} & \textbf{0.92} \tablehelper{0.07} & \textbf{0.92} \tablehelper{0.07} & 0.84 \tablehelper{0.11} & {0.81}\tablehelper{0.10} \\
%     500 & 0.63 \tablehelper{0.16} & \textbf{0.84} \tablehelper{0.21} & 0.45 \tablehelper{0.08} & 0.47  \tablehelper{0.09} & \textbf{0.82} \tablehelper{0.13} & \textbf{0.83} \tablehelper{0.13} & {0.43} \tablehelper{0.05} & {0.46}\tablehelper{0.11} \\
%     1000 & 0.43 \tablehelper{0.06} & \textbf{0.62} \tablehelper{0.12} & 0.45 \tablehelper{0.05} & 0.51  \tablehelper{0.05} & 0.61 \tablehelper{0.12} & \textbf{0.78} \tablehelper{0.16} & {0.42} \tablehelper{0.12} & {0.51}\tablehelper{0.20} \\
%     \bottomrule
% \end{tabular}
\begin{tabular}{c|cccc}
    \toprule
    \# other & MVAE & MoPoE-VAE & MMVAE & MMVAE++  \\ 
        \midrule
    50 & \textbf{0.96} \tablehelper{0.01} & \textbf{0.96}  \tablehelper{0.01} & \textbf{0.97} \tablehelper{0.01} & \textbf{0.96} \tablehelper{0.01}  \\
    250 & 0.72 \tablehelper{0.11} & 0.77  \tablehelper{0.07} & 0.76 \tablehelper{0.12} & \textbf{0.83} \tablehelper{0.09}  \\
    500 & 0.45 \tablehelper{0.08} & 0.47  \tablehelper{0.09} & 0.63 \tablehelper{0.16} & \textbf{0.84} \tablehelper{0.21}  \\
    1000 & 0.45 \tablehelper{0.05} & 0.51  \tablehelper{0.05} & 0.43 \tablehelper{0.06} & \textbf{0.62} \tablehelper{0.12}  \\
    \bottomrule \toprule
    & sup-MVAE & sup-MoPoE & sup-MMVAE & sup-MMVAE++  \\ 
        \midrule
    50 & \textbf{0.93} \tablehelper{0.04} & \textbf{0.93}\tablehelper{0.04} & \textbf{0.93} \tablehelper{0.04} & \textbf{0.93} \tablehelper{0.04}  \\
    250 & 0.84 \tablehelper{0.11} & {0.81}\tablehelper{0.10}& \textbf{0.92} \tablehelper{0.07} & \textbf{0.92} \tablehelper{0.07}  \\
    500 & {0.43} \tablehelper{0.05} & {0.46}\tablehelper{0.11}& \textbf{0.82} \tablehelper{0.13} & \textbf{0.83} \tablehelper{0.13}  \\
    1000 & {0.42} \tablehelper{0.12} & {0.51}\tablehelper{0.20}& 0.61 \tablehelper{0.12} & \textbf{0.78} \tablehelper{0.16}  \\
    \bottomrule
\end{tabular}
\end{sc}
\end{scriptsize}
\end{center}
\vskip -0.1in
\end{table*}

 \begin{figure}[!ht]
     \centering
     \includegraphics[width=\columnwidth]{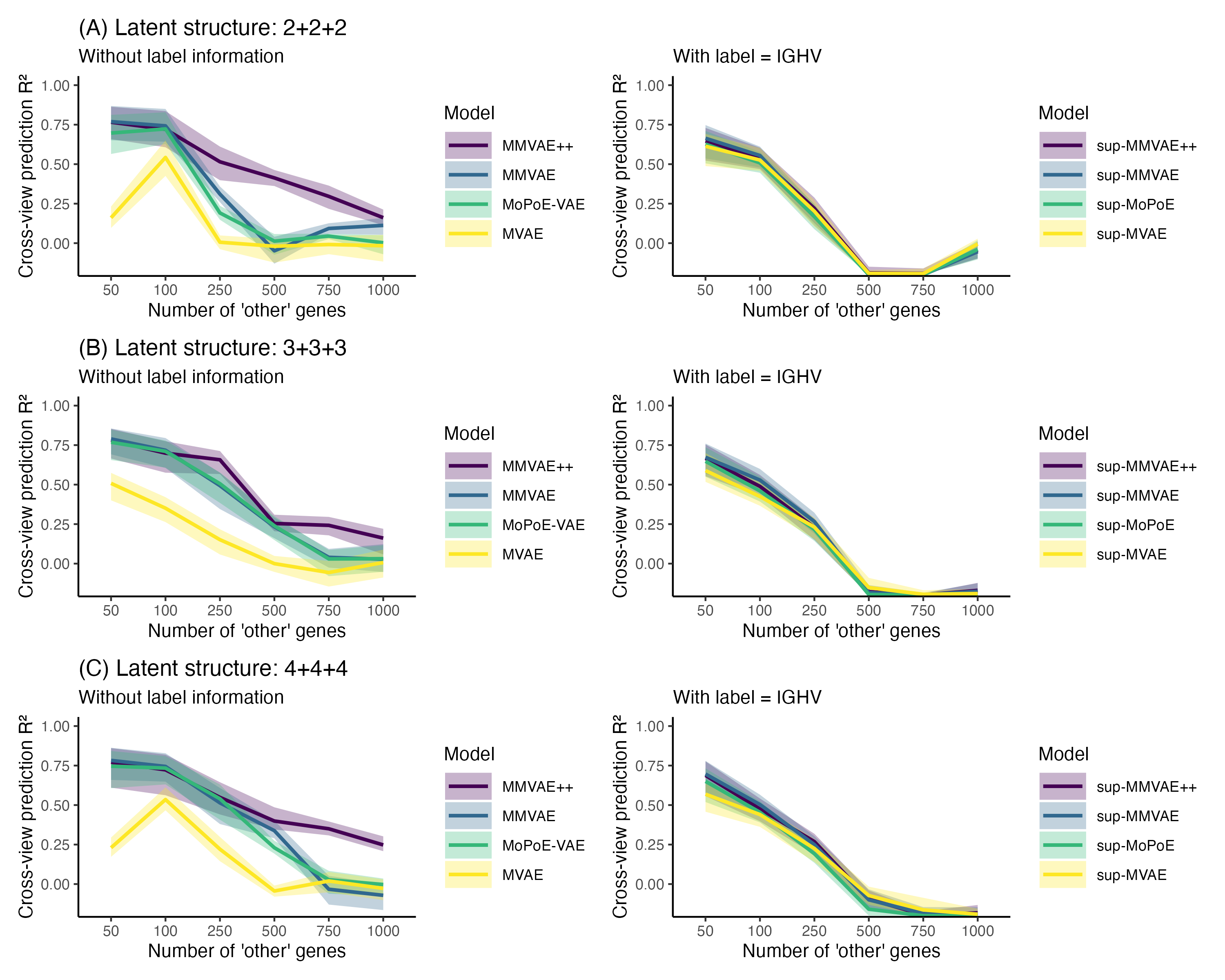}
     \caption{
       \textbf{CLL example expanded}, now showing also results for higher dimensional latent spaces, i.e.\ $[\text{dim}(\zprivateone), \text{dim}(\zshared),  \text{dim}(\zprivatetwo)]$ values in $[2, 2, 2], [3, 3, 3]$ and $[4, 4, 4]$. 
     }
     \label{fig:supp_CLL}
 \end{figure}

\newpage
\section{Single cell RNA + ATAC example: additional results}

\begin{figure}[!h]
    \centering
    \includegraphics[width=0.9\textwidth]{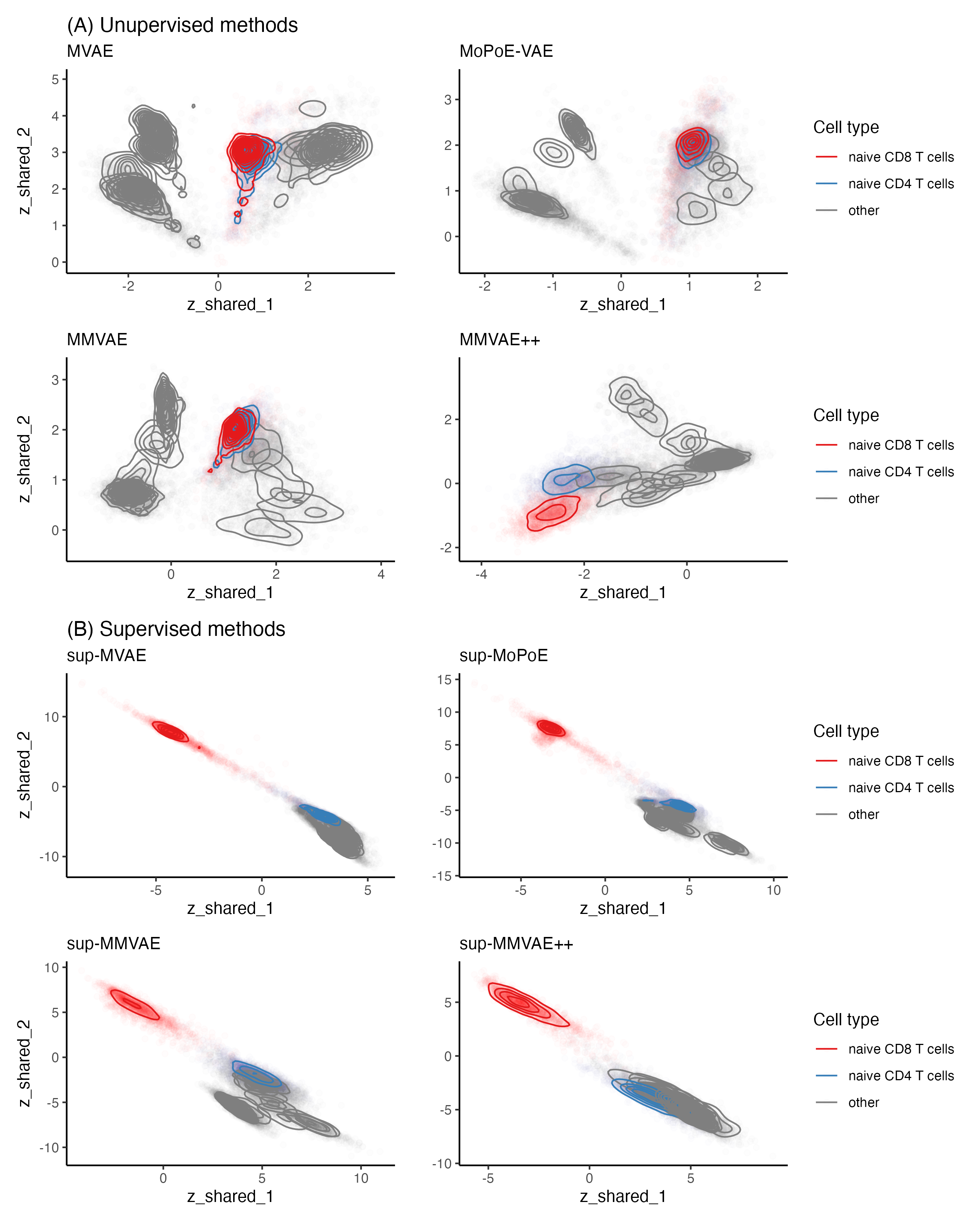}
    \caption{
    \textbf{Shared latent space visualisations in the single cell multi-omics example:} Highlighting the cell type of interest -- \textit{naive CD8 T cells} -- in red. Other cell types are shown in grey, but for the ease of interpretation, we have additionally highlighted naive \textit{CD4} T cells in blue, as we would expect it to be the closest among other cell types. Indeed, in the latent spaces inferred by MVAE, MoPoE-VAE and MMVAE, the blue and red cluster are highly overlapping, whereas MMVAE++ has learnt a latent space where CD8 and CD4 T cells are two distinct clusters. The separation is even more clear among supervised methods, but these have access to the class labels whereas MMVAE++ is an unsupervised approach. 
    }
    \label{fig:supp_single_cell}
\end{figure}

\end{document}